%% file: main.tex
\documentclass{article}
\usepackage{log_2024}						

\usepackage{booktabs}						
\usepackage{multirow}						
\usepackage{amsfonts}						
\usepackage{graphicx}						
\usepackage{duckuments}						
\usepackage{subfigure}
\usepackage{wrapfig,lipsum,booktabs}

\usepackage[numbers,compress,sort]{natbib}	

\newcommand{\first}[1]{\textcolor{red}{#1}}
\newcommand{\second}[1]{\textcolor{blue}{#1}}
\newcommand{\third}[1]{\textcolor{orange}{#1}}

\title[Cayley Graph Propagation]{Cayley Graph Propagation}

\author[JJ Wilson et al.]{%
JJ Wilson\\
Independent Researcher\\
\email{josephjwilson74@gmail.com}\And
Maya Bechler-Speicher \\
Tel-Aviv University\\
\email{mayab4@mail.tau.ac.il}\And
Petar Veli\v{c}kovi\'{c}\\
Google DeepMind\\
\email{petarv@google.com}
}

\begin{document}

\maketitle

\begin{abstract}
In spite of the plethora of success stories with graph neural networks (GNNs) on modelling graph-structured data, they are notoriously vulnerable to over-squashing, whereby tasks necessitate the mixing of information between distance pairs of nodes. To address this problem, prior work suggests rewiring the graph structure to improve information flow. Alternatively, a significant body of research has dedicated itself to discovering and precomputing bottleneck-free graph structures to ameliorate over-squashing. One well regarded family of bottleneck-free graphs within the mathematical community are \emph{expander graphs}, with prior work—Expander Graph Propagation (EGP)—proposing the use of a well-known expander graph family—the Cayley graphs of the $\mathrm{SL}(2,\mathbb{Z}_n)$ special linear group—as a computational template for GNNs. However, in EGP the computational graphs used are truncated to align with a given input graph. In this work, we show that truncation is detrimental to the coveted expansion properties. Instead, we propose CGP, a method to propagate information over a complete Cayley graph structure, thereby ensuring it is bottleneck-free to better alleviate over-squashing. Our empirical evidence across several real-world datasets not only shows that CGP recovers significant improvements as compared to EGP, but it is also akin to or outperforms computationally complex graph rewiring techniques.
\end{abstract}

\input{sections/introduction}

\section{Background}
\label{sec:background}
\input{sections/background}

\section{Existing approaches to mitigate over-squashing}
\input{sections/related_work}

\section{Benefits of Cayley graphs}
\label{sec:cayley_graph_benefits}
\input{sections/cayley_graph_benefits}

\section{Cayley Graph Propagation}
\label{sec:cgp}
\input{sections/cgp}

\section{Experimentation}
\label{sec:experimentation}
\input{sections/experimentation}

\section{Conclusion}
\input{sections/conclusion}

\clearpage

\section*{Acknowledgements}
We would like to thank Petar Veli\v{c}kovi\'{c}'s co-authors of Expander Graph Propagation, including Andreea Deac and Mark Lackenby. Their novel work of leveraging the use of expander graphs within the context of graph neural networks to alleviate bottlenecks and counter over-squashing provided the basis for our work. Furthermore, we would like to thank Alex Vitvitskyi, Csaba Szepesv\'ari, Johannes Vallikivi and Dobrik Georgiev for their invaluable insights prior to submission.

\bibliographystyle{unsrtnat}
\bibliography{reference}

\clearpage
\appendix

\section{Effective Resistance of Cayley graphs}
\label{app:effective_resistance}
\input{sections/appendices/effective_resistance}

\section{Virtual Nodes Initialisation}
\label{app:virtual_nodes}
\input{sections/appendices/virtual_nodes}

\section{Experimental Details}
\label{app:experimental_details}
\input{sections/appendices/experimental_details}

\section{Additional Experiments}
\label{app:additional_experiments}
\input{sections/appendices/additional_experiments}

\section{Scalability Analysis of CGP}
\label{app:scalability}
\input{sections/appendices/scalability}

\section{Cayley graphs as Regular graphs}
\label{app:regular_graphs}
\input{sections/appendices/regular_graphs}

\section{Dirichlet Energy of Cayley graphs}
\label{app:dirichlet_energy}
\input{sections/appendices/dirichlet_energy}

\end{document}

%% file: sections/introduction.tex
Graph neural networks (GNNs) have emerged as a cornerstone for processing graph-structured data~\cite{hamilton2017inductive} with significant contributions in various domains and real-world applications \cite{zhou2020graph, wu2020comprehensive}. The majority of GNNs are dependent on propagating information between neighbouring nodes in the graph \citep{bronstein2021geometric}, known as Message Passing Neural Networks (MPNNs) \citep{gilmer2017neural}. This \emph{message-passing} paradigm involves the iterative exchange of messages, with nodes aggregating and updating their representations based on received information from their neighbours. Thus, a sufficient number of layers are required to be able to capture long-range interactions. However, increasing the number of layers results in an exponential growth of the receptive field and subsequently leading to a large volume of messages to be aggregated into fixed-sized vectors. This phenomenon is known as \emph{over-squashing} \citep{alon2020bottleneck} and hinders the \emph{expressivity} of GNNs \cite{di2023over}. The underlying graph topology has been proven to be a key contributing factor to the over-squashing problem \cite{di2023does}.

The \emph{over-squashing} phenomenon is an active area of research with several techniques proposed to alter the topological properties of an input graph to mitigate over-squashing. Several recent works have analysed it through varying lenses, including curvature \cite{topping2022riccurvature}, spectral expansion properties \cite{karhadkar2023fosr, banerjee2022oversquashing} and effective resistance \cite{black2023understanding, arnaiz2022diffwire}. Most of the solutions to this problem fall into the category of \emph{graph rewiring}, in which the graph topology is directly modified based on an optimisation target. However, this imposes the computational complexity of having to surgically analyse the graph structure. Notably, the recent work of Expander Graph Propagation (EGP) by \citet{deac2022expander} propose a unique solution of constructing an independent desirable graph structure to propagate information over.

To this end, \citet{deac2022expander} identified four desirable criteria to mitigate over-squashing and effectively handle global context in graph representation learning: \emph{global information propagation} (i), \emph{no bottlenecks} (ii), \emph{subquadratic time and space complexity} (iii) and \emph{no dedicated preprocessing} (iv). The authors surveyed prior approaches, including traditional GNNs (iii, iv), master-node methods (i, iii, iv) \citep{gilmer2017neural, battaglia2018relational}, fully connected graphs (i, ii, iv) \citep{alon2020bottleneck} and the aforementioned graph rewiring techniques (i-iii). Ultimately, \citet{deac2022expander} recognised the efficacy of \emph{expander graphs} \cite{hoory2006expander, sarnak2004expander} as the desirable graph structure for bottleneck-free information propagation, due to their favourable topological properties. 

A family of expander graphs in \citet{deac2022expander} has been constructed leveraging the well-known theoretical results of \emph{special linear groups}, for which a family of corresponding Cayley graphs can be derived. This family of Cayley graphs is guaranteed to have the derisible topological properties to mitigate over-squashing, as well as being efficiently precomputable (i-iv). Although the constructed Cayley graphs are scalable, the number of nodes are in order of $O(|V|^3)$. This consideration is addressed within EGP \cite{deac2022expander} by identifying the smallest $n$ that yields a graph larger or equal to the desired number of nodes, and then subsequently \emph{truncating} the Cayley graph to match the input graph's number of nodes in breadth-first order. Consequently, this truncation procedure results in a \emph{subgraph} derived from the expander graph being used as the computational template.

Motivated by the promising research direction of \citet{deac2022expander} in which the authors use an independent graph structure that is theoretically known to exhibit desirable properties, we introduce an alternative approach of using the Cayley graph. We propose a more optimal approach of embracing the complete Cayley graph structure, guaranteeing the coveted topological properties.

\paragraph{Main contributions and outline}
In this paper, we present Cayley Graph Propagation (\textbf{CGP})\footnote{Our source code is available at:\url{https://github.com/josephjwilson/cayley_graph_propagation}}, a novel model that uses the complete Cayley graph structure to mitigate over-squashing, whilst still fulfilling the derisible criteria as set by \citet{deac2022expander} (i-iv). Our contributions are as follows:

\begin{itemize}
    \item In Section \ref{sec:cayley_graph_benefits}, 
    we highlight the optimal topological properties of Cayley graphs for message propagation. We show that the truncation procedure performed in EGP to align the Cayley graph with the input graph may be detrimental to the coveted theoretical benefits.
    \item In Section \ref{sec:cgp}, we introduce \textbf{CGP}, 
     a method to propagate information over the complete Cayley graph structure, thereby ensuring it is bottleneck-free and alleviates over-squashing. CGP modifies EGP by avoiding the truncation step used to align the input graph with a Cayley graph, utilising the additional nodes as virtual nodes.
    \item In Section \ref{sec:experimentation}, 
    we provide an empirical evaluation across several real-world datasets to show that CGP recovers significant improvements as compared to EGP. Additionally, our model is akin to or outperforms the computationally complex graph rewiring techniques.
\end{itemize}

%% file: sections/background.tex
\begin{figure}
    \centering
    \includegraphics[height=0.3\linewidth]{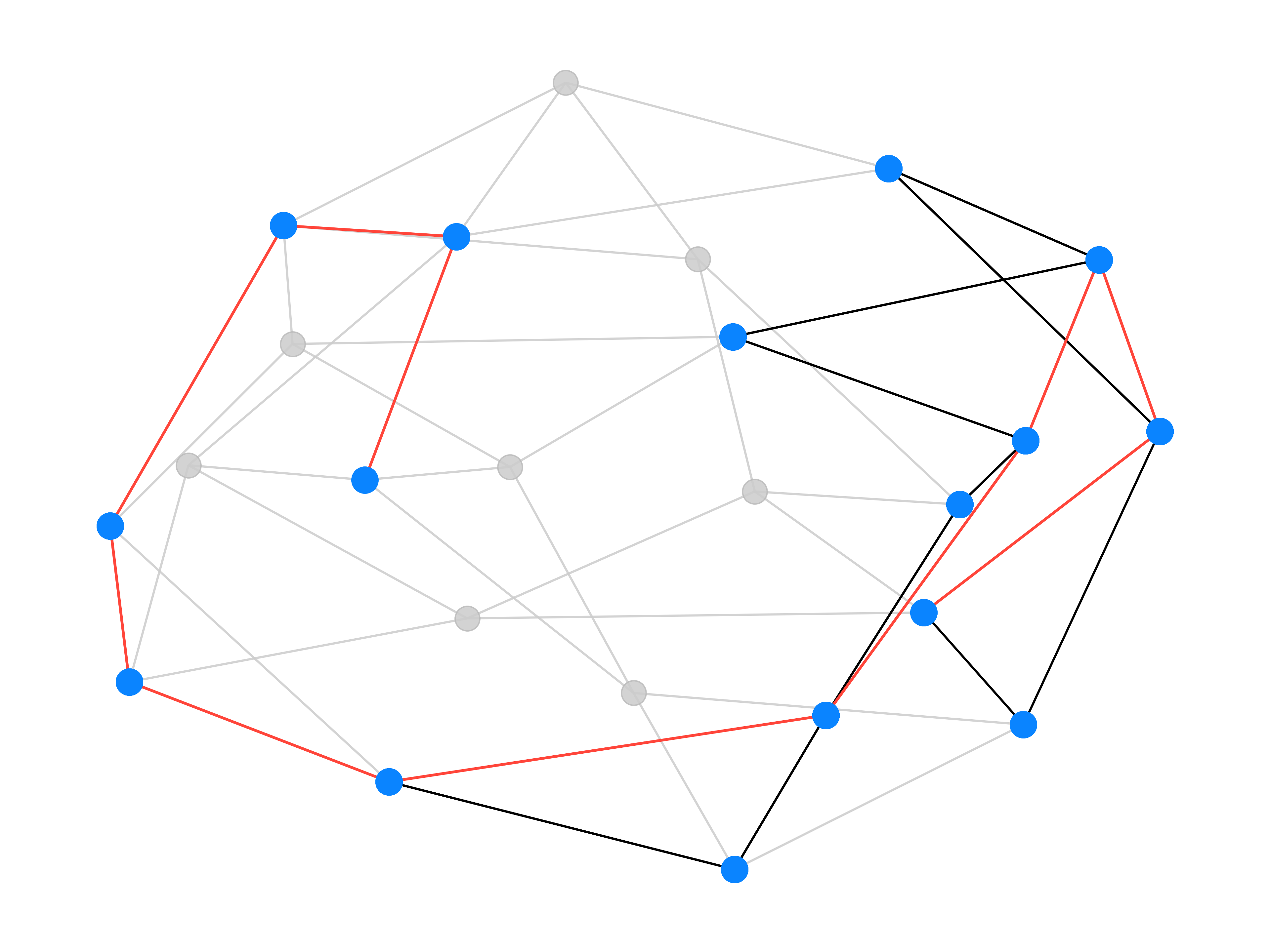}\hfill
    \includegraphics[height=0.3\linewidth]{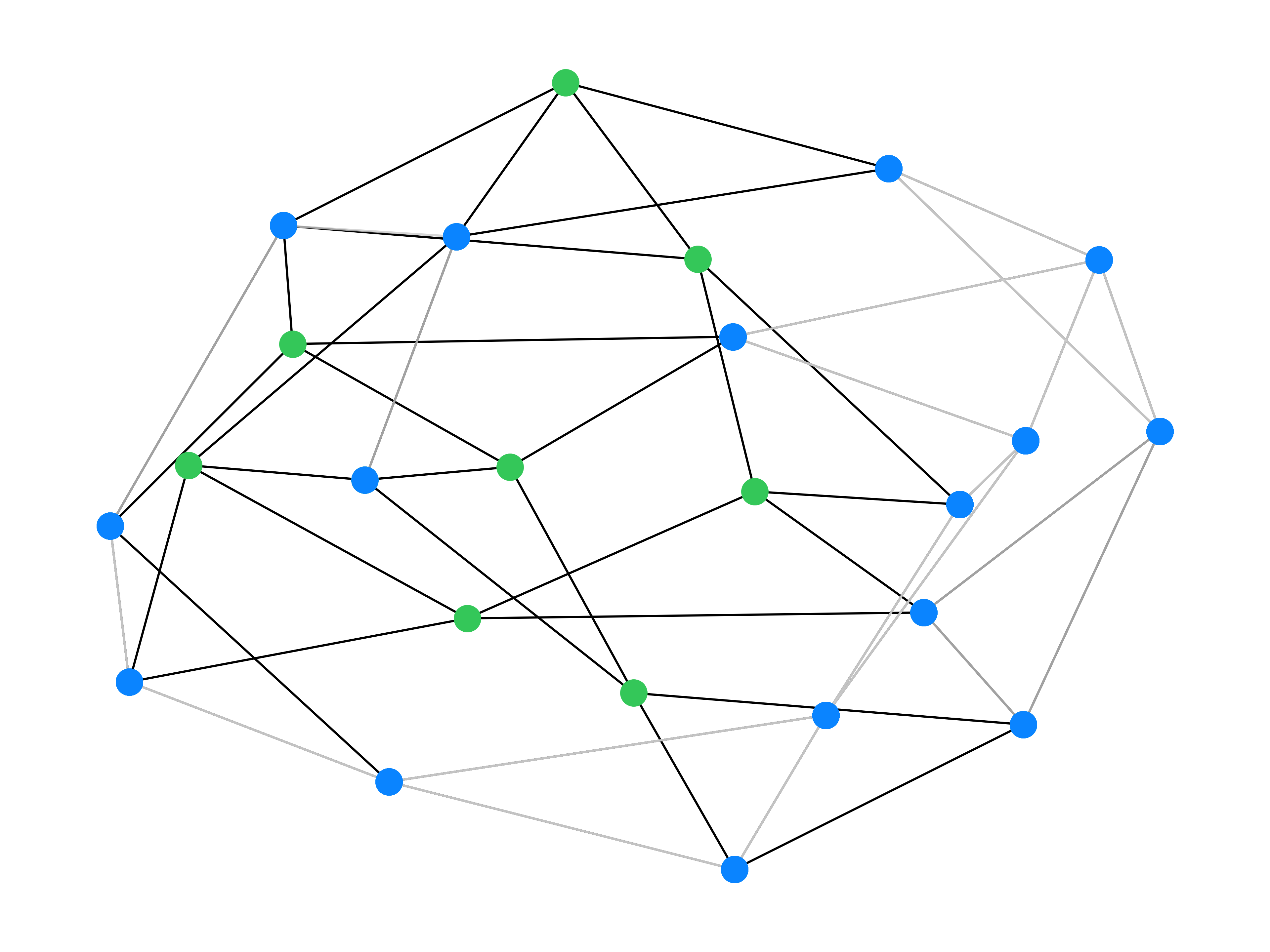}
    \caption{Both Cayley graphs represent $\mathrm{SL}(2, \mathbb{Z}_3)$ with $|V|=24$ nodes using the same construction. {\bf Left}: A truncated Cayley graph (\emph{spectral gap}: 0.0751, \emph{diameter}: 10) aligned to a given input graph. {\bf Right}: The \emph{complete} Cayley graph (\emph{spectral gap}: 1.2679, \emph{diameter}: 4) structure indicating the additional \emph{virtual nodes} (in green).}
    \label{fig:cayley_graph}
\end{figure}

\paragraph{Graph preliminaries}

Given an undirected graph denoted as $G = (V, E)$, where $V$ and $E$ denote its nodes and edges respectively. The topology of the graph is encoded in the adjacency matrix ${\bf A}\in\mathbb{R}^{|V| \times |V|}$, where $|V|$ is the number of nodes. Let ${\bf D} = {\bf D}(G)$ denote the diagonal matrix of degrees as given by ${\bf D}_{vv} = d_v$. The normalised graph Laplacian ${\bf L} = {\bf L}(G)$ is defined by ${\bf L} = {\bf D}^{-1/2}({\bf D} - {\bf A}){\bf D}^{-1/2}$. From the normalised Laplacian ${\bf L}$ the eigenvalues $0 = \lambda_0 \leq \lambda_1 \leq ... \leq \lambda_{n-2} \leq \lambda_{n-1}$. Importantly, from this derivation the \emph{spectral gap} of graph $G$ is $\lambda_1 - \lambda_0 = \lambda_1$; the \emph{Cheeger inequality} then defines that a larger spectral gap of graph $G$ is an indicator of good spectral expansion properties. Accordingly, a graph with desirable expansion properties (or a larger spectral gap) defines that it has strong connectivity, or alternatively it is globally lacking bottlenecks \cite{deac2022expander}. 

\paragraph{Expander graphs}
An expander graph is categorised by its unique properties of being both sparse and highly connected with the number of edges scaling linearly with the number of nodes $(|E| = O(|V|))$. One such expansion property that an expander graph satisfies is derived from the aforementioned Cheeger inequality. As a result, in essence, expander graphs do not have any bottlenecks \cite{banerjee2022oversquashing}; in Section~\ref{sec:cayley_graph_benefits} we further define and explore this link.

Due to the definition of an expander graph, there are consequently several known construction approaches \cite{kowalski2019introduction, davidoff2003elementary}. We will focus on the \emph{deterministic} algebraic approach as introduced by \citet{deac2022expander}. A family of expander graphs have been precomputed leveraging the well-known theoretical results of \emph{special linear groups}, $\mathrm{SL}(2,\mathbb{Z}_n)$, for which a family of corresponding Cayley graphs, $\mathrm{Cay}(\mathrm{SL}(2,\mathbb{Z}_n); S_n)$, can be derived. Here, $S_n$ (\cite{deac2022expander}, Definition 8) denotes a particular generating set for $\mathrm{SL}(2,\mathbb{Z}_n)$. For appropriate choices of $S_n$, the corresponding Cayley graphs are guaranteed to have expansion properties. Moreover, from Figure~\ref{fig:cayley_graph} we see that the constructed graph are 4-regular, $(|E| = 2|V|)$. Importantly, although Cayley graphs are scalable, achieving a specific number of nodes is not always feasible; for instance, the node count of $\mathrm{Cay}(\mathrm{SL}(2,\mathbb{Z}_n); S_n)$ is given as per:

\begin{equation}\label{eqn:caycount}
    \left|V(\mathrm{Cay}(\mathrm{SL}(2,\mathbb{Z}_n); S_n))\right| = n^3 \prod_{\text{prime } p|n} \left ( 1 - \frac{1}{p^2} \right ).
\end{equation}

\paragraph{Over-squashing}
The over-squashing problem was first identified by \citet{alon2020bottleneck}, whereby the information in a MPNN is aggregated from too many neighbours, meaning as a consequence they are squashed into fixed-size vectors. This can result in a loss of information \cite{shi2023exposition}. This phenomenon was then formalised \cite{topping2022riccurvature, black2023understanding, di2023over}, showing that the Jacobian of the node features is affected by topological properties of the graph, such as curvature and effective resistance. Furthermore, \citet{di2023does} analysed how over-squashing impacts the expressive power of GNNs. In the following section, we will address the literature and how current approaches aim to mitigate over-squashing.

%% file: sections/related_work.tex
In this section, we explore the current landscape of several novel techniques that try to alleviate the over-squashing phenomenon \cite{alon2020bottleneck}. In essence, the main principle behind many of these techniques is to decouple the input graph $G$ from the computational one, such that it has structurally fewer bottlenecks. \citet{alon2020bottleneck} simply proposed a rewiring technique that does not require the analysis of the input graph by making the last layer of the GNN fully adjacent (FA), allowing all nodes to interact with each other. The effectiveness of such an approach can be shown by (dense) Graph Transformers \cite{ying2021transformers, kreuzer2021rethinking}, where every layer is fully-connected. However, such an approach is limited by even modest graph sizes due to it imposing $O(|V|^2)$ edges. An alternative approach is a master node \cite{gilmer2017neural}; here a new node is introduced, which is connected to all of the nodes within the graph. This approach is effective as it reduces the graph's diameter to $2$ by only adding one new node with $O(|V|)$ edges. However, the master node itself becomes the bottleneck. Notably, both of the aforementioned approaches are independent in relation to the input graph topology, therefore satisfying (iv) of \emph{no dedicated preprocessing}.

\paragraph{Graph rewiring}
An alternative promising approach line of research is to \emph{rewire} the input graph $G$ to optimise the \emph{spectral} or \emph{spatial} properties of the graph. To this end, an abundance of graph rewiring techniques have stemmed to modify the graph connectivity to try and mitigate bottlenecks. A popular class of approaches are based on a \emph{spectral} quantity of a graph \cite{karhadkar2023fosr} or to reduce the effective resistance \cite{arnaiz2022diffwire, banerjee2022oversquashing, black2023understanding}. These approaches have provided promising insights and have empirically reduced over-squashing, but they impose a computational complexity of having to examine the input graph structure.

\paragraph{Expander graph based rewiring}
The existing approaches of quantitative analysis leads to an alternative approach being the understanding of the desirable graph structure known as an expander graph. An expander graph exhibits the desirable properties associated with \emph{spectral gap} and \emph{effective resistance}. For this reason, \citet{banerjee2022oversquashing} proposes a construction inspired by an expander graph to randomly locally rewire a given input graph, whilst \citet{shirzad2023exphormer} use both \emph{virtual global nodes} and expander graphs as a powerful primitive to design a more scalable graph transformer architecture. As previously examined, the work of \citet{deac2022expander} proposes a different schema of precomputing a bank of expander graphs, which are then interwoven by alternating layers on the input graph and then the auxiliary expander layer. This schema has proven to also be successful in high-order structures~\cite{christie2023higher} and in the first rewiring approach on temporal graphs~\cite{petrovic2024temporal}.

%% file: sections/cayley_graph_benefits.tex
\begin{figure}
    \includegraphics[width=0.45\linewidth]{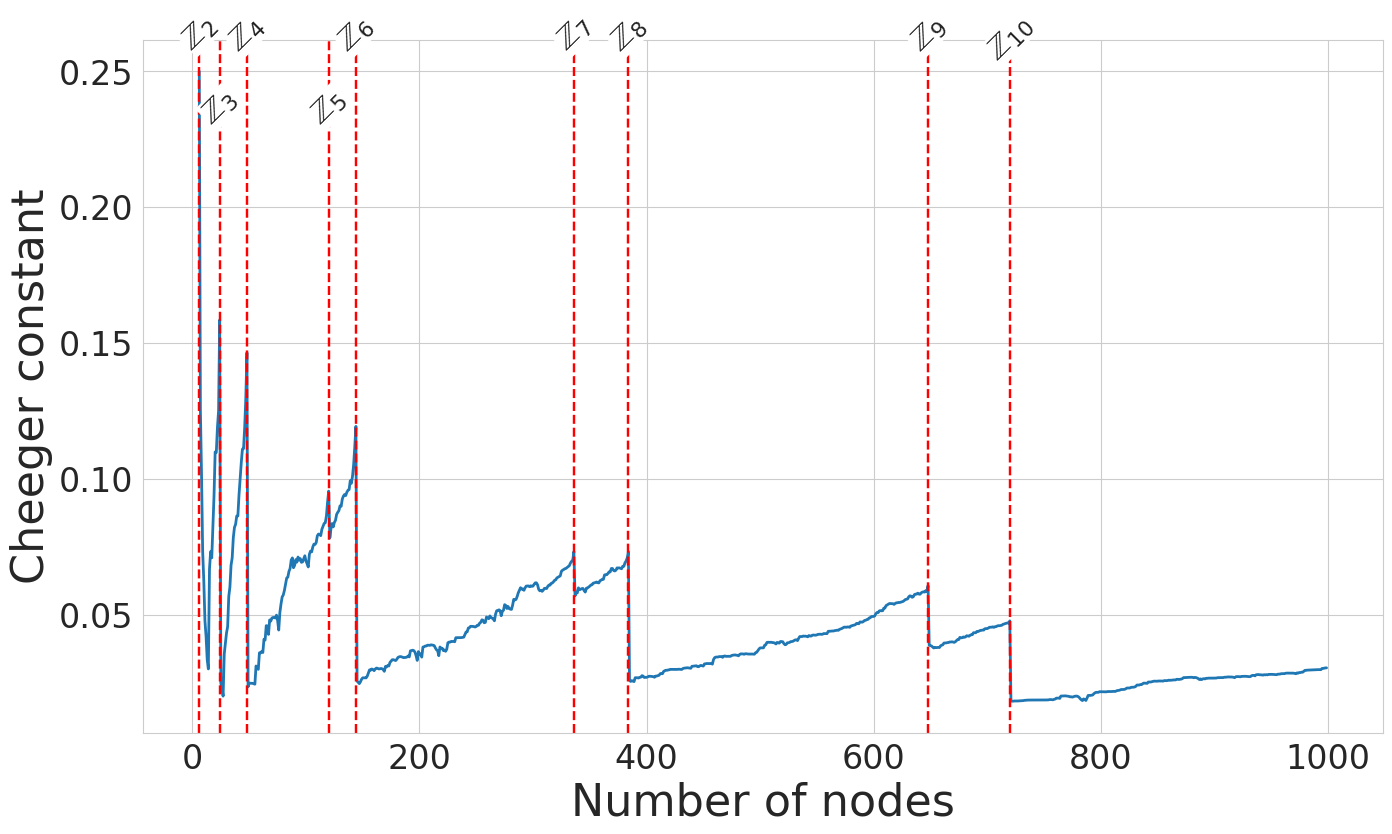}
     \hfill
    \includegraphics[width=0.45\linewidth]{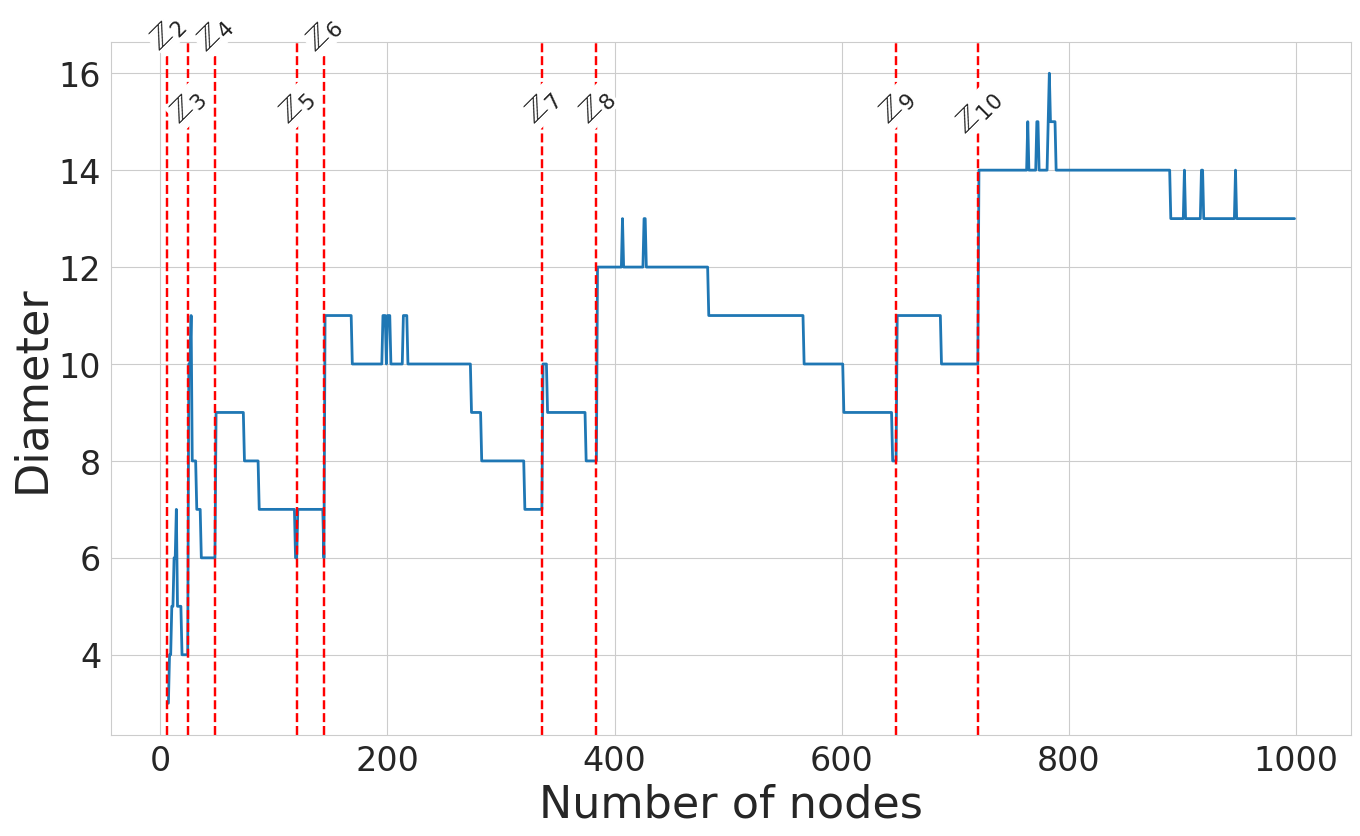}
    \caption{Illustrates the correlation between the Cheeger constant (left) and diameter (right) between extracting a truncated Cayley graphs from $\mathrm{SL}(2,\mathbb{Z}_n)$. The properties exhibit the most derisible property at each given complete Cayley graph interval (as denoted by the dotted red line) for the groups of $\mathbb{Z}_n$. The trend shows a detrimental drop in the derisible property occurring once the truncated Cayley graph aligns with the next complete Cayley graph structure.}
    \label{fig:expansion_properties}
\end{figure}

In this section, we formalise the topological properties of expander graphs and why they have been used as a conduit in a number of over-squashing approaches \cite{banerjee2022oversquashing, shirzad2023exphormer, deac2022expander}. In particular, we focus on the Cayley graph expander family as constructed by \citet{deac2022expander}. Furthermore, we importantly anaylse the impact on the topological properties of the truncated Cayley graphs as used by \citet{deac2022expander}. In contrast to this, in the appendix, we examine an alternative benefit to Cayley graphs being regular graphs through the lens of the recent work of \citet{bechler2023graph}.

\subsection{Topological properties of Cayley graphs}
The Cayley graph's topology serves as ideal conduit for the propagation of information due to its sparsity and being highly connected. In particular, we recall that the family of Cayley graphs derived from $\mathrm{Cay}(\mathrm{SL}(2,\mathbb{Z}_n); S_n)$ are in the magnitude of $O(|V|^3)$, and are well-known to exhibit a high spectral gap, low diameter and optimal commute time.

\paragraph{Connectivity}
One metric to measure a graph's connectivity is the coveted \emph{Cheeger constant} \cite{chungbook}. It provides a measurement of the narrowest bottleneck in a graph; a higher Cheeger constant indicates that a graph is globally lacking bottlenecks. Alternatively, it effectively describes how difficult it is to separate a graph $G$ into two subgraphs by removing edges. The exact Cheeger constant $h(G)$ is known to be a computationally challenging problem. To address this difficulty, we recall the \emph{spectral gap} of a graph $G$ (defined in Section \ref{sec:background}), which provides a bound for the Cheeger constant from discrete Cheeger inequality \cite{cheeger1970lower, alon1984eigenvalues}. As per \citet{chungbook}, this bound is:

\begin{equation}\label{eqn:cheeger_inequality}
\frac{\lambda_1}{2} \leq h(G) \leq \sqrt{2 \lambda_1},
\end{equation}

where $\lambda_1$ is the second-smallest eigenvalue or spectral gap of the normalised graph Laplacian ${\bf L}(G)$. In Figure~\ref{fig:expansion_properties}, we use the lower bound from the Cheeger constant; the figure illustrates that the complete Cayley graph structure exhibits the most desirable Cheeger constant (or higher spectral gap). In turn, it is observed that the most unfavourable scenario occurs for a truncated Cayley graph just beyond the range of the proceeding graph, as derived from the \emph{special linear group} $\mathrm{SL}(2,\mathbb{Z}_n)$.

\paragraph{Diameter}
The diameter of a graph influences the effectiveness of traversal between nodes. Intuitively, a lower diameter facilitates a more efficient graph structure, enabling nodes to reach each other in a shorter number of hops. Our constructed Cayley graph with $|V|$ nodes has a low diameter, requiring only $O(log(|V(G_i)|)$ (\cite{deac2022expander}, Theorem 5) steps to globally propagate information. The results in Figure \ref{fig:expansion_properties} in relation to the diameter correlate with those shown for the Cheeger constant; the lower diameter is at each complete Cayley graph structure.

\paragraph{Commute time}
The recent work of \citet{di2023over, di2023does} validates that the topology of the graph is the primary factor affecting over-squashing. Theorem 5.5 from \citet{di2023over} establishes that the extent of over-squashing between a pair of nodes $u$ and $v$ over-squashing can be bounded by the commute time $T(u,v)$. The commute time is defined as the expected number of steps in a random walk from $u$ to $v$ and return back to $u$. Consequently, over-squashing occurs between nodes with a large commute time. Notably, the family of Cayley graphs as derived by \citet{deac2022expander} are directly mentioned by \citet{di2023over, di2023does} as being the \emph{optimal} computational template for message-passing due to the commute time scaling linearly with the number of edges. We note that the commute-time $T(u,v)$ between a pair of nodes is analogous to the \emph{effective resistance} \cite{black2023understanding}. In Appendix (Section \ref{app:effective_resistance}), we extend our analysis of the impact of truncating a Cayley graph to align with an input graph through the perspective of effective resistance.

%% file: sections/cgp.tex
In the previous sections, we provided theoretical motivations for our proposed method of utilising the complete Cayley graph structure. The setup for CGP closely aligns with that of EGP in most aspects. We consider the input to a GNN as a node feature matrix ${\bf X}\in\mathbb{R}^{|V|\times d}$ and an adjacency matrix ${\bf A}\in\mathbb{R}^{|V|\times|V|}$.

The construction of the Cayley graph $\mathrm{Cay}(\mathrm{SL}(2, \mathbb{Z}_n); S_n)$ is done by choosing the smallest $n$ such that $|V(\mathrm{Cay}(\mathrm{SL}(2, \mathbb{Z}_n); S_n))|\geq|V|$. However, we no longer truncate the Cayley graph such that a subgraph ${\bf A}^{\text{Cay}(n)}_{1:|V|,1:|V|}$ is extracted -- instead, we opt for a different approach of retaining all of the nodes of the Cayley graph, and its corresponding adjacency matrix ${\bf A}^{\text{Cay}(n)}$.

This construction requires us to add new nodes into the graph; hence, we need to modify the feature matrix into an extended version, ${\bf X}^{\text{Cay}(n)}\in\mathbb{R}^{|V(\mathrm{Cay}(n))|\times d}$. To construct this, we featurise the first $|V|$ nodes using the data from ${\bf X}$, and treat any additional nodes as \emph{virtual nodes}, initialised in some pre-defined way. Specifically:
\begin{equation}
    {\bf X}^{\text{Cay}(n)}_{1:|V|} = {\bf X} \qquad \qquad {\bf X}^{\text{Cay}(n)}_{|V|+1:|V(\text{Cay}(n)|} \sim \mathrm{InitVirt}
\end{equation}
where $\mathrm{InitVirt}$ is any sampling procedure for initialising $d$-dimensional feature vectors. For example, we may choose to sample random features from $\mathcal{N}(0, 1)$, or in our case initialise them to zeros \cite{gilmer2017neural, cai2023connection, southern2024understanding}. In Appendix (Section \ref{app:virtual_nodes}), we provide additional studies to compare the performance of different initialisation strategies.

Because EGP makes advantage of both the input graph (specified by ${\bf A}$) and the generated Cayley graph (specified by ${\bf A}^{\mathrm{Cay}(n)}$), we can also appropriately extend the original adjacency matrix, ${\bf A}$, to incorporate the new nodes. Since the input graph layers are intended to preserve the input graph topology as much as possible, one approach is to construct such a matrix $\widetilde{\bf A}\in\mathbb{R}^{|V(\mathrm{Cay}(n)|\times|V(\mathrm{Cay}(n)|}$ by adding \emph{self-edges} to the virtual nodes only:
\begin{equation}
    \widetilde{\bf A}_{1:|V|,1:|V|} = {\bf A} \qquad \qquad \widetilde{\bf A}_{|V|+1:|V(\text{Cay}(n)|,|V|+1:|V(\text{Cay}(n)|} = {\bf I}
\end{equation}
\begin{equation}
\widetilde{\bf A}_{1:|V|,|V|+1:|V(\text{Cay}(n)|} = \widetilde{\bf A}_{|V|+1:|V(\text{Cay}(n)|, 1:|V|} = {\bf 0}
\end{equation}
CGP now proceeds in the same manner as EGP: alternating GNN layers, such that every odd layer operates over the input graph---to preserve the topological information therein---and every even layer operates over the generated Cayley graph---to support bottleneck-free global communication. For a two-layer CGP model, this can be depicted as:
\begin{equation}
    {\bf H} = \mathrm{GNN}(\mathrm{GNN}({\bf X}^{\mathrm{Cay}(n)}, \widetilde{\bf A}; {\theta}_1), {\bf A}^{\mathrm{Cay}(n)}; {\theta}_2)
\end{equation}
where $\theta_1$ and $\theta_2$ are the parameters of the first and second GNN layer, respectively. This implementation works with any choice of base GNN; here we may choose to take advantage of the graph isomorphism network (\cite{xu2018powerful}, GIN):
\begin{equation}\label{eqn:gin}
    \vec{h}_u = \phi\left(\left(1 + \epsilon\right)\vec{x}_u + \sum_{v\in\mathcal{N}_u} \vec{x}_v\right)
\end{equation}
where $\vec{x}_u\in\mathbb{R}^d$ are the features of node $u$, $\epsilon$ is a learnable scalar, and $\phi$ is a MLP. Our experimentation in Section \ref{sec:experimentation} shows that CGP is MPNN agnostic. 

The final node embeddings ${\bf H}\in\mathbb{R}^{|V(\mathrm{Cay}(n)|\times k}$ may then be used for downstream node, graph or graph-level tasks. To avoid direct influence of virtual nodes in these predictions, we use only the embeddings corresponding to the original graph's nodes, that is, ${\bf H}_{1:|V|}$, in downstream tasks.

The CGP model upholds the requirements of the four criteria (i-iv) set by \citet{deac2022expander}---arguably, in a more theoretically grounded way than EGP; Figure \ref{fig:cayley_graph} and \ref{fig:expansion_properties} provides empirical evidence of this. Specifically, the lower diameter of the graph used in CGP enhances its ability to eliminate over-squashing and bottlenecks, which is further supported by having a higher spectral gap. Furthermore, the CGP model may be able to make up for one of the limitations of the Cayley graph construction: the inability to find the best way to align it to a given input graph, mitigating the potential for \emph{stochastic} effects in the process. The additional virtual nodes act as ``bridges'' between poorly connected communities in the Cayley graph, ameliorating any poorly-connected regions caused by misalignment.

%% file: sections/experimentation.tex
In this section, we empirically validate the efficacy of using the complete Cayley graph structure on a range of graph classification tasks. In particular, we first motivate CGP by extending the results of \citet{deac2022expander}, comparing CGP against approaches that do not incur a computational complexity of having to examine the input graph's structure: master node \cite{gilmer2017neural}, fully adjacent last layer (FA) \cite{alon2020bottleneck} and EGP \cite{deac2022expander}. We then extend our empirical evaluation by comparing the performance of CGP against the state-of-the-art approaches that require \emph{dedicated preprocessing}~\cite{gasteiger2019digl, topping2022riccurvature, karhadkar2023fosr, nguyen2023revisiting, black2023understanding,choipanda} on the TUDataset~\cite{morris2020tudataset} and LRGB~\cite{dwivedi2022long}. Due to space limitations, we defer the results of the LRGB to Appendix (Section \ref{app:additional_experiments}).
 
Beyond comparing the performance of CGP against the baselines, we evaluate the practicality of CGP by providing a runtime analysis against the aforementioned approaches that require dedicated processing. In Appendix~(Section \ref{app:scalability}), we extend this through a scalability analysis. Furthermore, we provide an ablation study to investigate whether the complete Cayley graph structure is a suitable alternative to a fully-connected graph.
 
\begin{wraptable}{r}{8cm}
    \vspace{-5mm}
    \small
    \centering
    \caption{Comparative performance evaluation of CGP against the baselines on the OGB. OOM denotes out-of-memory on a \textsc{NVIDIA RTX 4090}.}\label{tab:ogb}
    \begin{tabular}{l ccc}\toprule
        \multirow{2}{*}{Model} &\multicolumn{1}{c}{\textsc{ogbg-molhiv}} &\multicolumn{1}{c}{\textsc{ogbg-ppa}}  \\ \cmidrule(lr){2-2} \cmidrule(lr){3-3}
        &\textbf{Test ROC-AUC $\uparrow$} & \textbf{Test ACC $\uparrow$} \\\midrule
        \midrule
        GCN & $0.7566 \pm 0.0104$ & $0.5483 \pm 0.0209$ \\
        \;+ Master Node & $0.7531 \pm 0.0128$ & $0.5824 \pm 0.0219$ \\
        \;+ FA & $0.7628 \pm 0.0191$  & OOM \\
        \;+ EGP & $0.7731 \pm 0.0081$ & $\textbf{0.6821} \pm 0.0045$ \\
        \;+ \textbf{CGP} & $\textbf{0.7794} \pm 0.0122$ & $0.6782 \pm 0.0066$ \\
        \midrule \midrule
        GIN & $0.7678 \pm 0.0183$ & $0.5888 \pm 0.0441$ \\
        \;+ Master Node & $0.7608 \pm 0.0134$ & $0.6069 \pm 0.0062$ \\
        \;+ FA & $0.7718 \pm 0.0147$  &  OOM \\
        \;+ EGP & $0.7537 \pm 0.0076$ &  $0.6533 \pm 0.119$ \\
        \;+ \textbf{CGP} & $\textbf{0.7899} \pm 0.0090$ & $\textbf{0.6562} \pm 0.0147$ \\
        \bottomrule
    \end{tabular}
    \vspace{-2mm}
\end{wraptable}

\paragraph{Experimental setting}
We evaluate on the Open Graph Benchmark (OGB) \cite{hu2020open} and TUDataset \cite{morris2020tudataset}. In our experiments, we prioritise a fair comparison, following the layer schema for each approach. For EGP and CGP, this includes the interweaving schema as depicted in Section \ref{sec:cgp}. For FA \cite{alon2020bottleneck}, this consists of rewiring the last layer. All the other approaches only propagate over the base (or rewired) graph structure. Furthermore, we show that our proposed method is MPNN invariant by setting the underlying model to GCN~\cite{kipf2016semi} and GIN~\cite{xu2018powerful}. The chosen hyperparameters are in line with established foundations \cite{hu2020open, karhadkar2023fosr} for each dataset respectively. Notably, EGP \cite{deac2022expander} limits its empirical analysis to only the graph classification tasks from OGB with GIN as the backbone GNN. Refer to Appendix (Section \ref{app:experimental_details}) for more details on our experimental setting and hyperparameters used for the OGB and TUDataset.

\paragraph{OGB}
For a real-world comparison and to extend the foundations of EGP, we first provide results on two graph classification tasks, \textsc{ogbg-molhiv} and  \textsc{ogbg-ppa}, from the OGB \cite{hu2020open}. We compare CGP against techniques that do not require \emph{dedicated preprocessing}. \textsc{ogbg-molhiv} is among the largest molecule property prediction datasets within the scope of the MoleculeNet benchmark \citep{wu2018moleculenet}, thus providing emulation for real-world analysis. \textsc{ogbg-ppa} focuses on classifying species based on their taxa, using their protein-protein association networks \cite{proteins}. Our model takes inspiration from the open-source implementation of OGB and hyperparameters as given by \cite{hu2020open}, including fixing the number of layers to 5, a hidden dimension of 300, a dropout of 50\% and with the only modification being a batch size of 64. We report across 10 seeds and 5 seeds for \textsc{ogbg-molhiv} and \textsc{ogbg-ppa} respectively. Our results in Table~\ref{tab:ogb} show that overall the CGP model outperforms the other approaches that do not require \emph{dedicated preprocessing}; exemplified in the results for \textsc{ogbg-molhiv}. Moreover, CGP consistently outperforms the base GCN and GIN, which is not the case for the other baseline models.

\begin{table}[t!]
    \scriptsize
    \setlength{\tabcolsep}{3pt}
    \centering
    \caption{Results of CGP compared against EGP, FA and the approaches that require dedicated preprocessing for GCN and GIN on the TUDataset. The experimental setup uses the setting as in \citet{karhadkar2023fosr}, and hyperparameters for each baseline from \cite{gasteiger2019digl, topping2022riccurvature, karhadkar2023fosr, nguyen2023revisiting, black2023understanding, choipanda}. The colours highlight \first{First}, \second{Second} and \third{Third}. OOT indicates out-of-time for the \emph{dedicated prepreprocessing} time.}
    \begin{tabular}{l cccccc}\toprule
        Model & \textsc{REDDIT-BINARY} & \textsc{IMDB-BINARY} & \textsc{MUTAG} & \textsc{ENZYMES} & \textsc{PROTEINS} & \textsc{COLLAB} \\ \midrule
        GCN & $77.735 \pm 1.586$ & $\third{60.500} \pm 2.729$ & $74.750 \pm 4.030$ & $29.083 \pm 2.363$ & $66.652 \pm 1.933$ & $70.490 \pm 1.628$ \\
        \;+ FA & OOM & $48.950 \pm 1.652$ & $70.250 \pm 4.608$ & $28.667 \pm 3.693$ & $71.071 \pm 1.506$ & $\third{72.039} \pm 0.771$ \\
        \;+ DIGL & $77.350 \pm 1.206$ & $49.600 \pm 2.435$ & $70.500 \pm 5.045$ & $\second{30.833} \pm 1.537$ & $72.723 \pm 1.420$ & $56.470 \pm 0.865$ \\ 
        \;+ SDRF & $\third{77.975} \pm 1.479$ & $59.000 \pm 2.254$ & $74.000 \pm 3.462$ & $26.667 \pm 2.000$ & $67.277 \pm 2.170$ & $71.330 \pm 0.807$ \\ 
        \;+ FoSR & $77.750 \pm 1.385$ & $59.750 \pm 2.357$ & $75.250 \pm 5.722$ & $24.167 \pm 3.005$ & $70.848 \pm 1.618$ & $67.220 \pm 1.367$ \\
        \;+ BORF & OOT & $48.900 \pm 0.900$ & $\second{76.750} \pm 0.037$ & $27.833 \pm 0.029$ & $67.411 \pm 0.016$ & OOT \\ 
        \;+ GTR & $\second{79.025} \pm 1.248$ & $\second{60.700} \pm 2.079$ & $76.500 \pm 4.189$ & $25.333 \pm 2.931$ & $\third{72.991} \pm 1.956$ & $\second{72.600} \pm 1.025$ \\
        \;+ PANDA & $\first{87.275} \pm 1.033$ & $\first{68.350} \pm 2.346$ & $\third{76.750} \pm 5.531$ & $\third{30.667} \pm 2.019$ & $70.134 \pm 1.518$ & $\first{73.850} \pm 0.695$  \\
        \;+ EGP & $67.550 \pm 1.200$ & $59.700 \pm 2.371$ & $70.500 \pm 4.738$ & $27.583 \pm 3.262$ & $\first{73.304} \pm 2.516$ & $69.470 \pm 0.970$ \\
        \midrule
        \;+ CGP & $67.050 \pm 1.483$ & $56.200 \pm 1.825$ & $\first{83.750} \pm 3.597$ & $\first{31.000} \pm 2.397$ & $\second{73.036} \pm 1.291$ & $69.630 \pm 0.730$\\
        \midrule \midrule
        GIN & $84.600 \pm 1.454$ & $\second{71.250} \pm 1.509$ & $80.500 \pm 5.143$ & $35.667 \pm 2.803$ & $70.312 \pm 1.749$ & $71.490 \pm 0.746$ \\
        \;+ FA & OOM & $69.900 \pm 2.332$ & $80.250 \pm 5.314$ & $\second{47.833} \pm 2.529$ & $\second{72.902} \pm 1.419$ & $72.740 \pm 0.786$ \\
        \;+ DIGL & $84.575 \pm 1.265$ & $52.650 \pm 2.150$ & $78.500 \pm 4.189$ & $41.500 \pm 3.063$ & $\third{72.321} \pm 1.440$ & $57.620 \pm 1.010$ \\
        \;+ SDRF & $84.550 \pm 1.396$ & $69.550 \pm 2.381$ & $80.500 \pm 4.177$ & $37.167 \pm 2.709$ & $69.509 \pm 1.709$ & $\third{72.958} \pm 0.419$ \\
        \;+ FoSR & $\second{85.750} \pm 1.099$ & $69.250 \pm 1.810$ & $80.500 \pm 4.738$ & $28.083 \pm 2.301$ & $71.518 \pm 1.767$ & $71.720 \pm 0.892$ \\
        \;+ BORF & OOT & $\third{70.700} \pm 0.018$ & $79.250 \pm 0.038$ & $34.167 \pm 0.029$ & $70.625 \pm 0.017$ & OOT \\
        \;+ GTR & $\third{85.474} \pm 0.826$ & $69.550 \pm 1.473$ & $79.000 \pm 3.847$ & $31.750 \pm 2.466$ & $72.054 \pm 1.510$ & $71.849 \pm 0.710 $  \\
        \;+ PANDA & $\first{90.325} \pm 0.867$ & $68.350 \pm 2.346$ & $\second{83.250} \pm 3.262$ & $\third{42.167} \pm 2.286$ & $72.321 \pm 1.786$ & $\second{73.320} \pm 0.814 $  \\
        \;+ EGP & $77.875 \pm 1.563$ & $68.250 \pm 1.121$ & $\third{81.500} \pm 4.696$ & $40.667 \pm 3.095$ & $70.848 \pm 1.568$ & $72.330 \pm 0.954$\\
        \midrule
        \;+ CGP & $78.225 \pm 1.268$ & $\first{71.650} \pm 1.532$ & $\first{85.250} \pm 3.200$ & $\first{50.083} \pm 2.242$ & $\first{73.080} \pm 1.396$ & $\first{73.350} \pm 0.788$ \\
        \bottomrule
    \end{tabular}
    \label{tab:graph_rewiring}
\end{table}

\paragraph{TUDataset}
We extend our graph classification task analysis by evaluating on \textsc{REDDIT-BINARY}, \textsc{IMDB-BINARY}, \textsc{MUTAG}, \textsc{ENZYMES}, \textsc{PROTEINS} and \textsc{COLLAB} from the TUDataset \cite{morris2020tudataset}. Significantly, these datasets were chosen under the claim of \citet{karhadkar2023fosr} that the topology of the graphs in relation to the tasks require long-range interactions. Accordingly, the TUDataset has become the cornerstone collection of benchmark datasets for the \emph{graph-rewiring} approaches investigating over-squashing. Thus, we compare CGP against EGP \cite{deac2022expander} and FA \cite{alon1984eigenvalues}, as well as the following state-of-the-art graph rewiring techniques that require \emph{dedicated preprocessing}: DIGL \cite{gasteiger2019digl}, SDRF \cite{topping2022riccurvature}, FoSR \cite{karhadkar2023fosr}, BORF \cite{nguyen2023revisiting} and GTR \cite{black2023understanding}. Moreover, we include PANDA \cite{choipanda} as a unique approach that dynamically alters the \emph{width} \cite{di2023over} of the model to alleviate \emph{over-squashing} as opposed to rewiring the input graph structure.

The results in Table~\ref{tab:graph_rewiring} prioritise a fair comparison to further pinpoint with certainty that the performance gain can be credited to the utilisation of the complete Cayley graph structure. Therefore, in Table~\ref{tab:graph_rewiring} we use the hyperparameters and experimental setting as prescribed by \citet{karhadkar2023fosr}. The hyperparameters include a hidden dimension of 64, the number of layers set to 4 and a dropout of 50\%. In line with our OGB experimentation for all models, we also use Batch Norm~\cite{ioffe2015batch}. Unlike the baseline model, FA, EGP and CGP, the \emph{dedicated preprocessing} approaches have an optimisation target and thus feature additional approach specific hyperparameters. For each dataset, we use the optimised hyperparameter setting as stated in each paper respectively. More details are found in Appendix (Section \ref{app:experimental_details}). The reported results are averaged across 20 random seeds and \citet{karhadkar2023fosr} reports the results with a 95\% confidence interval, thus we respect this for the TUDataset. Notably, we report OOT to indicate out-of-time for the \emph{preprocessing} rewiring procedure for BORF on the \textsc{REDDIT-BINARY} and \textsc{COLLAB} datasets. This is in accordance with \citet{nguyen2023revisiting}, which do not report results for these two datasets and corresponding hyperparameters. In addition, a time-out is reported in \cite{choipanda} for the aforementioned datasets, whilst in \cite{barbero2023locality} they report out-of-memory for \textsc{COLLAB}.

The results in Table~\ref{tab:graph_rewiring} underscore the effectiveness of CGP in comparison with state-of-the-art baselines. In particular, in the case of GIN, our CGP model obtains the highest accuracy for all datasets except for \textsc{REDDIT-BINARY}. The overall performance of CGP when applied to GCN is not as competitive as those of GIN, however our results are still comparable with the other baselines. 
This is particularly notable when the sparsity of the Cayley graphs is considered in relation to certain datasets, such as \textsc{IMDB-BINARY} and \textsc{COLLAB}. The sparse nature of the Cayley graph
means that unlike many graph rewiring techniques edges may be removed; refer to Appendix (Section \ref{app:experimental_details}) for the dataset statistics. However, the results of CGP for GIN recover this loss in performance for \textsc{IMDB-BINARY} and \textsc{COLLAB}, emphasising the work of \citet{you2020design} in which the design space of GNNs can greatly impact the results of a model. Finally, of significance is the parity of the hyperparameter's number of layers; Table~\ref{tab:ogb} uses 5 layers, whereas Table~\ref{tab:graph_rewiring} uses 4 layers. This demonstrates the performance of CGP is irrespective of the final layer being the input or Cayley graph.

\begin{table}[t!]
    \scriptsize
    \centering
    \caption{CGP training, evaluation time (seconds per epoch), and memory consumption statistics in comparison to baseline models on \textsc{REDDIT-BINARY} and \textsc{COLLAB} from the TUDataset~\cite{morris2020tudataset}. OOM signifies out-of-memory on a \textsc{NVIDIA RTX 4090}.}
    \label{tab:tu_complexity}
    \begin{tabular}{l ccc ccc}
        \toprule
        \multirow{2}{*}{Model} & \multicolumn{3}{c}{\textsc{REDDIT-BINARY}} & \multicolumn{3}{c}{\textsc{COLLAB}} \\ 
        \cmidrule(lr){2-4} \cmidrule(lr){5-7} 
        & \textbf{Train Time} & \textbf{Eval. Time} & \textbf{Mem. (GB)} 
        & \textbf{Train Time} & \textbf{Eval. Time} & \textbf{Mem. (GB)} \\
        \midrule
        GIN & $0.1049 \pm 0.0237$ & $0.0741 \pm 0.0032$ & $922$ & $0.2787 \pm 0.0345$ & $0.2364 \pm 0.0094$ & $1722$ \\
        \;+ FA & OOM & OOM & OOM & $0.4625 \pm 0.0507$ & $0.4488 \pm 0.0404$ & $4746$ \\
        \;+ FoSR & $0.1117 \pm 0.0268$ & $0.0841 \pm 0.0164$ & $906$ & $0.3129 \pm 0.0386$ & $0.2619 \pm 0.0257$ & $3320$ \\
        \;+ PANDA & $0.7902 \pm 0.0597$ & $0.7489 \pm 0.0439$ & $1316$ & $2.1152 \pm 0.0964$ & $1.9347 \pm 0.0924$ & $4406$ \\
        \;+ EGP & $0.1215 \pm 0.0257$ & $0.0952 \pm 0.0128$ & $976$ & $0.3096 \pm 0.0372$ & $0.2598 \pm 0.0164$ & $1696$ \\
        \midrule
        \;+ \textbf{CGP} & $0.1326 \pm 0.0296$ & $0.1147 \pm 0.0135$ & $1128$ & $0.3191 \pm 0.0321$ & $0.2785 \pm 0.0160$ & $2418$ \\
        \bottomrule
    \end{tabular}
\end{table}

\paragraph{Scalability}
In the following, we investigate whether the \emph{additional virtual nodes} as leveraged in CGP will introduce an increased \emph{computational complexity}, impacting the training time of the model. Accordingly, we compare CGP directly against EGP \cite{deac2022expander}, FA \cite{alon2020bottleneck} and the graph-rewiring approach FoSR \cite{karhadkar2023fosr}. Moreover, we include PANDA to provide extra detail on the results found in Table \ref{tab:graph_rewiring}. Even though PANDA obtains state-of-the-art performance, as stated by \citet{choipanda} this approach impacts the runtime. We choose \textsc{REDDIT-BINARY} and \textsc{COLLAB} from the TUDataset because they have the largest average graph sizes among the collection of datasets. In Table~\ref{tab:tu_complexity} we report the average seconds per epoch for the training and evaluation time, as well as the memory consumption statistics, using GIN as the underlying model. 

The results highlight that \emph{additional virtual nodes} in CGP have a negligible impact on the training and evaluation time. Even though many virtual nodes may be added, they are sparsely connected. This is depicted in Figure \ref{fig:cayley_graph}, where the 4-regular structure of the Cayley graph results in virtual nodes being sparsely connected compared to other virtual node approaches, such as a master node~\cite{gilmer2017neural}. The base model, FoSR \cite{karhadkar2023fosr}, EGP \cite{deac2022expander} and CGP are shown to be akin with each other. CGP is shown to increase the memory consumption, however the sparsely connected virtual nodes have a minimal impact when compared with FA \cite{alon1984eigenvalues}, FoSR \cite{karhadkar2023fosr} and PANDA \cite{choipanda}. The dataset statistics in Appendix (Section \ref{app:experimental_details}) provide an explanation, such as \textsc{REDDIT-BINARY} being several times larger than all other datasets found in the TUDataset \cite{morris2020tudataset}. Overall, the results in Table~\ref{tab:tu_complexity} highlight the strengths of CGP against the approaches that require dedicated preprocessing, as well as the dense fully adjacent layer \cite{alon1984eigenvalues}. In Appendix (Section \ref{app:scalability}), we extend our scalability analysis of CGP. This includes conducting a scalability analysis to compare the \emph{dedicated preprocessing} time of the techniques reported in Table~\ref{tab:graph_rewiring} on the real-world datasets from the TUDataset, as well as further extending this evaluation through a synthetic benchmark as used by \citet{karhadkar2023fosr}.

\paragraph{Ablation studies}
In the following, we answer the question: \emph{`is the complete Cayley graph structure a suitable alternative to a fully adjacent layer \cite{alon2020bottleneck}?'}. In Table~\ref{tab:tu_ablations}, we show that it is a promising avenue as the results for CGP$^{\dagger}$ and FA$^{\dagger}$ are similar for both GCN and GIN.  However, the sparsity of CGP is highlighted in the results of \textsc{REDDIT-BINARY}, whereby FA runs out-of-memory (OOM). Accordingly, these results demonstrate the advantages of CGP, due to it being far more scalable. Notably, the results reported in Table~\ref{tab:tu_ablations} use the same experimental setting and hyperparameters as Table~\ref{tab:graph_rewiring}.

Next, we investigate in accordance with the recent work of \citet{bechlergraph}; we examine if we can ignore the input graph entirely and solely propagate over the Cayley graph structure. Our results in Table~\ref{tab:tu_ablations} suggest that the inductive bias endowed from the input graph still is required. The baseline procedure of CGP (interweaving a Cayley graph with an input graph), and using a Cayley graph for the last layer only, outperforms using a Cayley graph solely for each layer. One explanation is that the interweaving schema of CGP aligns with the principles of JK networks \cite{xu2018representation}, facilitating improved structure-aware representations by varying the neighbourhood ranges. However, the tone set by \citet{bechlergraph} is still a promising line of research, as the results indicate that the optimal graph-structure used is still task dependent.

\begin{table}[t!]
    \scriptsize
    \setlength{\tabcolsep}{3pt}
    \centering
    \caption{Results of CGP using the Cayley graph in different layer approaches compared against FA on the TUDataset. $\dagger$ denotes last layer and $\ast$ denotes every layer. OOM is out-of-memory on a NVIDIA RTX 4090. The colours highlight \first{First}, \second{Second} and \third{Third}.}
    \begin{tabular}{l cccccc}\toprule
        Model & \textsc{REDDIT-BINARY} & \textsc{IMDB-BINARY} & \textsc{MUTAG} & \textsc{ENZYMES} & \textsc{PROTEINS} & \textsc{COLLAB} \\ \midrule
        GCN & $\first{77.735} \pm 1.586$ & $\first{60.500} \pm 2.729$ & $74.750 \pm 4.030$ & $\third{29.083} \pm 2.363$ & $66.652 \pm 1.933$ & $\second{70.490} \pm 1.628$ \\
        \;+ FA$^{\dagger}$ & OOM & $48.950 \pm 1.652$ & $70.250 \pm 4.608$ & $28.667 \pm 3.693$ & $71.071 \pm 1.506$ & $\first{72.039} \pm 0.771$ \\
        \;+ FA$^{\ast}$ & OOM & $49.700 \pm 1.871$ & $75.250 \pm 4.554$ & $27.167 \pm 2.770$ & $\second{71.384} \pm 1.380$ & $54.990 \pm 0.699$ \\
        \midrule
        \;+ CGP & $\second{67.050} \pm 1.483$ & $\second{56.200} \pm 1.825$ & $\first{83.750} \pm 3.597$ & $\first{31.000} \pm 2.397$ & $\first{73.036} \pm 1.291$ & $\third{69.630} \pm 0.730$\\
        \;+ CGP$^{\dagger}$ & $\third{56.025} \pm 2.185$ & $51.200 \pm 1.729$ & $\second{77.750} \pm 3.706$ & $\second{30.167} \pm 3.196$ & $66.875 \pm 2.441$ & $59.050 \pm 1.077$ \\
        \;+ CGP$^{\ast}$ & $54.949 \pm 1.531$ & $\third{52.350} \pm 2.242$ & $\third{76.000} \pm 4.219$ & $28.918 \pm 3.191$ & $\third{68.124} \pm 3.346$ & $58.450 \pm 1.321$ \\
        \midrule \midrule
        GIN & $\first{84.600} \pm 1.454$ & $71.250 \pm 1.509$ & $80.500 \pm 5.143$ & $35.667 \pm 2.803$ & $70.312 \pm 1.749$ & $71.490 \pm 0.746$ \\
        \;+ FA$^{\dagger}$ & OOM & $69.900 \pm 2.332$ & $80.250 \pm 5.314$ & $47.833 \pm 2.529$ & $\second{72.902} \pm 1.419$ & $\third{72.740} \pm 0.786$ \\
        \;+ FA$^{\ast}$ & OOM & $54.250 \pm 4.784$ & $83.750 \pm 7.224$ & $34.250 \pm 4.669$ & $71.250 \pm 4.721$ & $55.270 \pm 1.604$ \\
        \midrule
        \;+ CGP & $78.225 \pm 1.268$ & $\first{71.650} \pm 1.532$ & $\third{85.250} \pm 3.200$ & $\second{50.083} \pm 2.242$ & $\first{73.080} \pm 1.396$ & $\first{73.350} \pm 0.788$ \\
        \;+ CGP$^{\dagger}$ & $\second{82.700} \pm 2.474$ & $\second{71.500} \pm 1.646$ & $\first{88.500} \pm 4.955$ & $\first{52.000} \pm 3.355$ & $\third{71.383} \pm 2.472$ & $\second{72.880} \pm 0.834$ \\
        \;+ CGP$^{\ast}$ & $\third{80.925} \pm 1.976$ & $\third{71.400} \pm 2.002$ & $\second{85.500} \pm 3.309$ & $\third{49.167} \pm 3.184$ & $70.937 \pm 1.788$ & $72.340 \pm 1.038$ \\
        \bottomrule
    \end{tabular}
    \label{tab:tu_ablations}
\end{table}

%% file: sections/conclusion.tex
In this work, we presented Cayley Graph Propagation (\textbf{CGP}), an efficient propagation scheme that mitigates over-squashing. CGP utilises the complete Cayley Graph structure to guarantee improved information flow between nodes in the input graph. We highlight the advantageous topological properties of Cayley graphs for message-passing. We show that by truncating the Cayley graphs to align with the input graph, as suggested in Expander Graph Propagation, the resulting graph may contain bottlenecks. This is in contrast to the Cayley graph we use, which is guaranteed to be bottleneck-free. We demonstrate the effectiveness and efficiency of CGP compared to EGP and other rewiring approaches, over multiple real-world datasets, including large-scale and long-range datasets.

\paragraph{Limitations and Future Work}
One limitation of our proposed model is the performance on datasets containing graphs with a comparatively higher node-to-edge ratio. For this reason, one such avenue for future work is aligning the Cayley graph edges such that that they retain the inductive bias of the task  \cite{sterner2024commute}. Additionally, it would be interesting to see how CGP performs in other tasks that utilise expander graphs, including but not limited to temporal graph rewiring \cite{petrovic2024temporal}. Furthermore, concurrent work has used \emph{virtual nodes} as the focal point of their proposed methods \cite{qian2024probabilistic, sestak2024vn} with \citet{southern2024understanding} analysing the role of virtual nodes within the context of over-squashing. Therefore, we theorise an interesting setting would be applying these authors' approaches to the additional virtual nodes retained from the complete Cayley graph structure.

%% file: sections/appendices/effective_resistance.tex
\begin{figure}[t!]
    \centering
    \subfigure[MUTAG]{\includegraphics[width=0.32\textwidth]{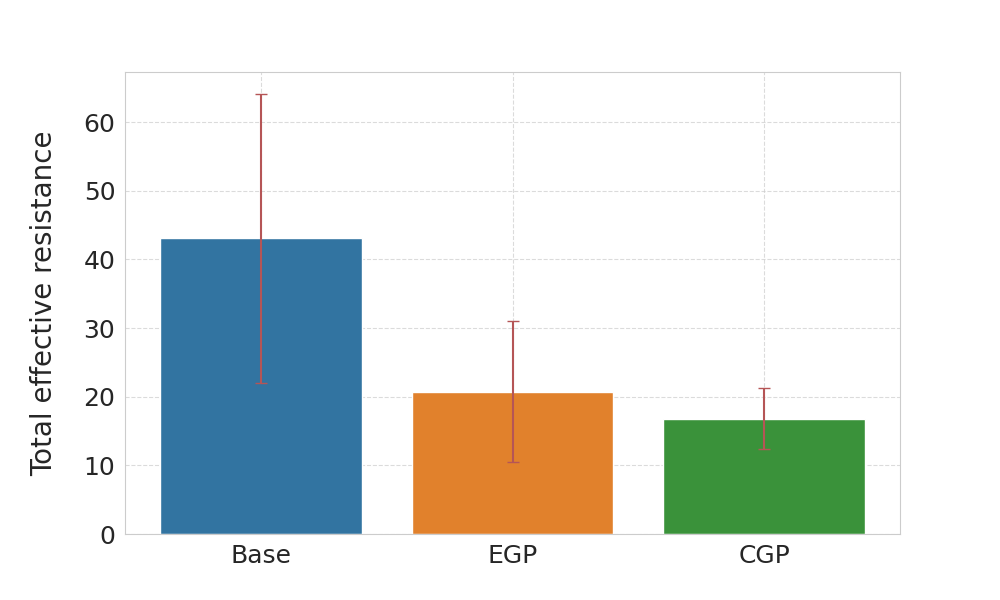}} 
    \subfigure[ENZYMES]{\includegraphics[width=0.32\textwidth]{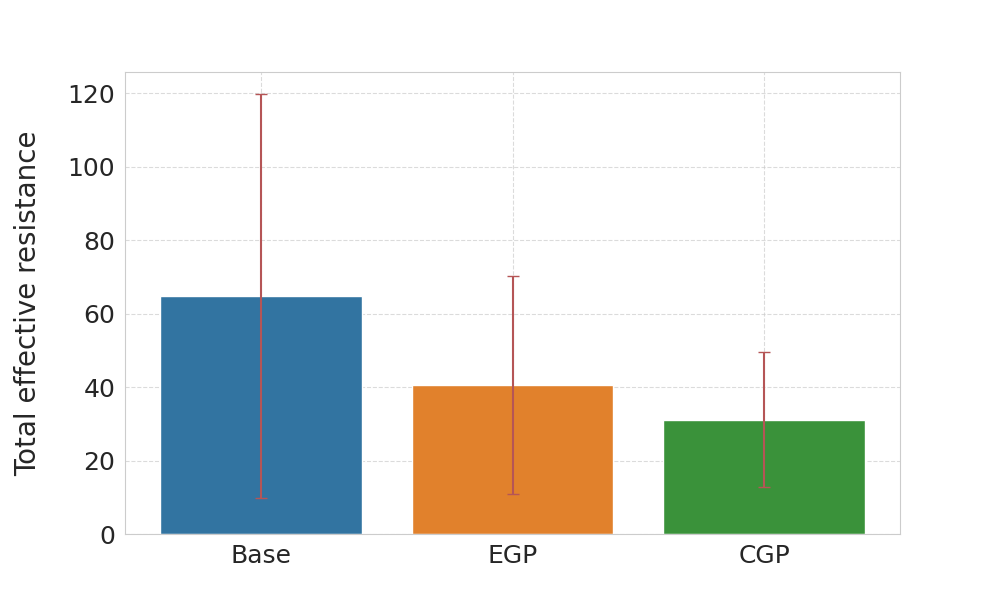}}
    \subfigure[PROTEINS]{\includegraphics[width=0.32\textwidth]{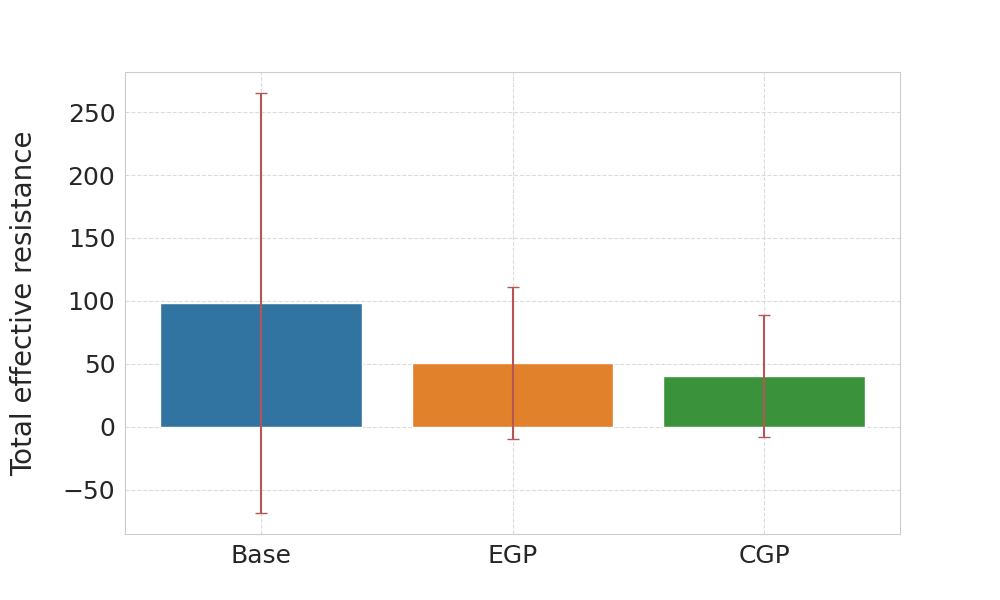}}
    \subfigure[IMDB-BINARY]{\includegraphics[width=0.32\textwidth]{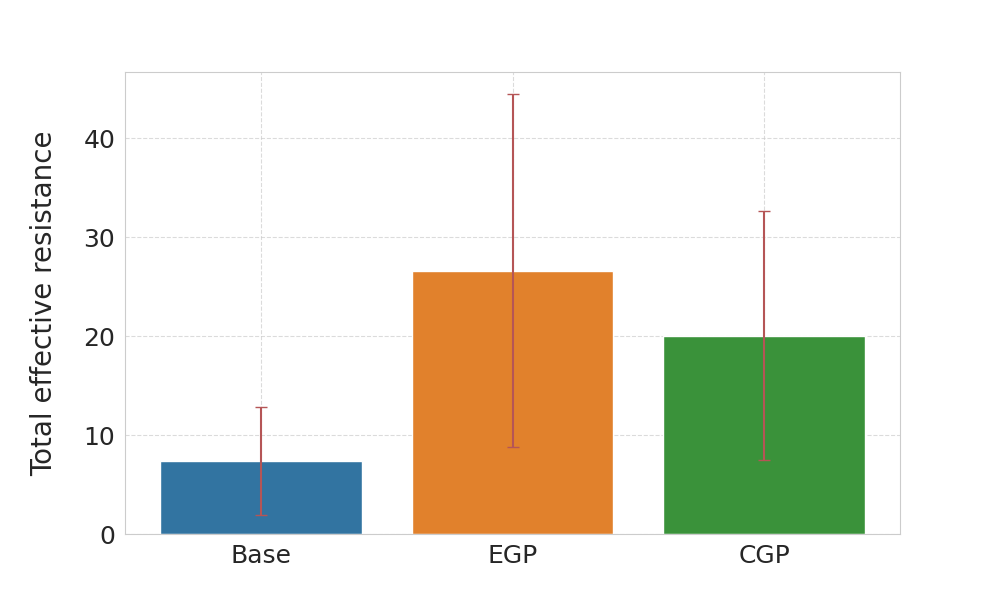}} 
    \subfigure[COLLAB]{\includegraphics[width=0.32\textwidth]{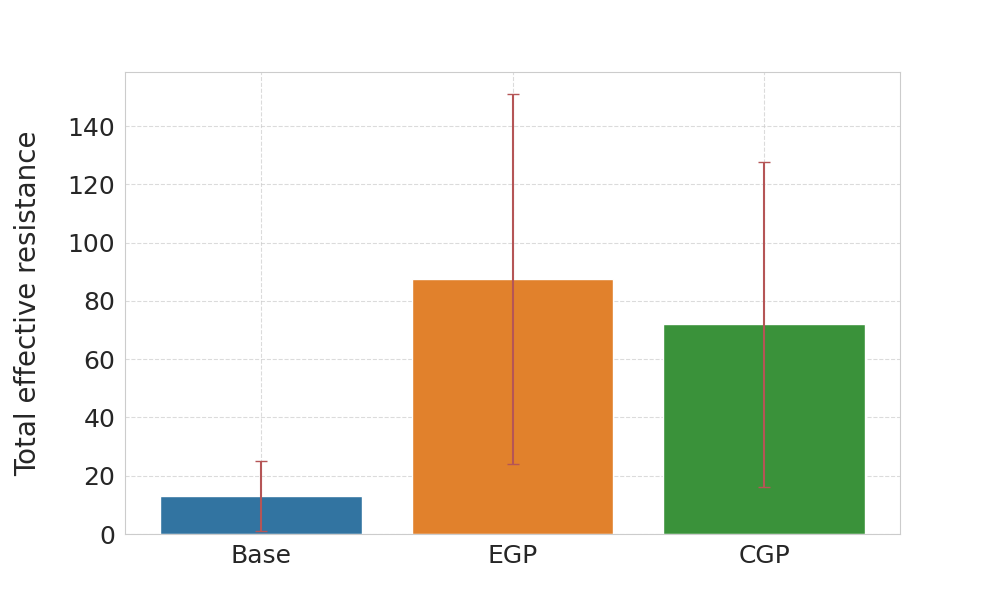}}
    \subfigure[OGBG-MOLHIV]{\includegraphics[width=0.32\textwidth]{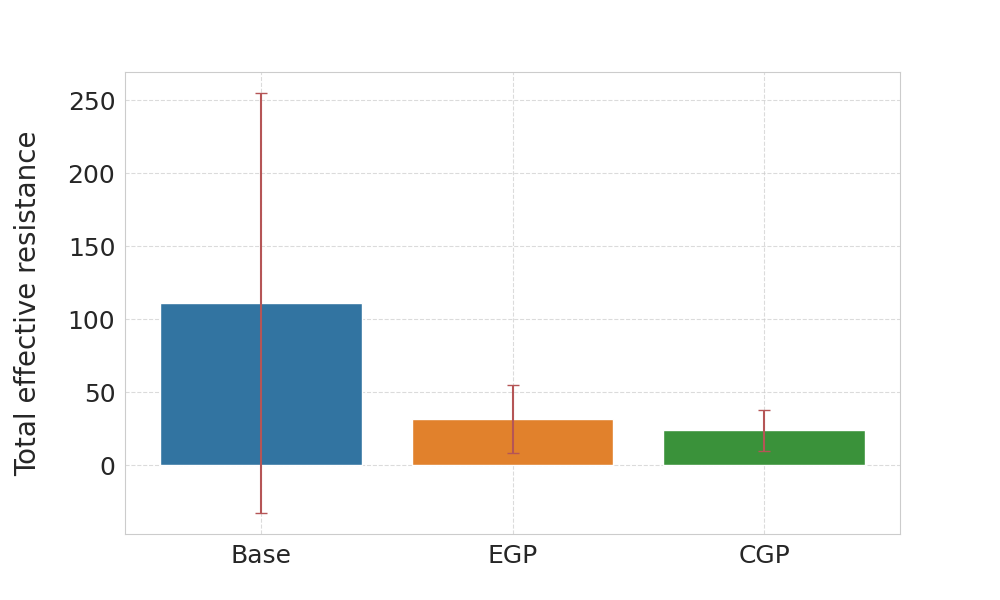}}
    \caption{Comparison of the total effective resistance $R_{tot}$ for CGP against the baseline model and EGP. A lower total effective resistance indicates that a graph is less susceptible to over-squashing.}
    \label{fig:effective_resistance}
\end{figure}

In this work, we have used the \emph{Cheeger constant} as an approach to measure \emph{bottlenecks} in a graph \cite{memoli2022persistent} in regards to \emph{over-squashing}. An alternative closely related approach is measuring \emph{over-squashing} through the lens of \emph{effective resistance} \cite{black2023understanding}. Stemming from the field of electrical engineering, the effective resistance between two nodes $u$ and $v$ reflects the ease of current flow. In turn, this concept has become analogous to measuring the connectivity between nodes within graph theory. Formally, the effective resistance between two nodes $u$ and $v$ can be expressed using the pseudoinverse ${\bf L}^{+}$ of $\mathbf{L}$, $R_{u,v} = ({\bf 1}_u - {\bf 1}_v)^{\mathsf{T}} {\bf L}^{+}({\bf 1}_u - {\bf 1}_v)$, where ${\bf 1}_u$ and ${\bf 1}_v$ are indicators for nodes $u$ and $v$.

The \emph{total effective resistance} $R_{tot}$ then builds upon this by measuring the \emph{effective resistance} for all pairs of nodes within a graph, thus providing a metric to quantify \emph{over-squashing} in a graph. As per \citet{black2023understanding}, the \emph{total effective resistance} $R_{tot}$ is given by:

\begin{equation}
    R_{tot} \,=\, \sum_{u>v} R_{u,v} \,=\, n\cdot \mathrm{Tr}(\mathbf{L}^{+}) \,=\, n\sum_{i=1}^{n-1}\frac{1}{\lambda_{i}}.\label{eq:r_total}
\end{equation}

The results in Figure~\ref{fig:effective_resistance} show the average of the \emph{total effective resistance} $R_{tot}$ for all the corresponding Cayley graphs against the base input graphs and truncated Cayley graphs as found in EGP. Akin to the results presented in \citet{black2023understanding}, for a fair evaluation we do not include graphs that may be disconnected. This is because the Cayley graphs used in CGP are a complete graph structure, therefore every node will be connected. The results show that CGP consistently has a lower total effective resistance $R_{tot}$ in comparison to EGP. Significantly, the complete Cayley graph structure is chosen by recalling $|V(\mathrm{Cay}(\mathrm{SL}(2, \mathbb{Z}_n); S_n))|\geq|V|$, therefore the total effective resistance $R_{tot}$ for CGP may be inflated due to it potentially being summed over more pairs of nodes. This further reinforces our claim that it is more beneficial to use the complete Cayley graph structure with the \emph{additional nodes} serving as shortcuts for message passing between nodes along the graph.

In line with our results reported in our empirical evaluation, for certain datasets the total effective resistance for EGP and CGP is higher than the base input graph. The statistics of the datasets reported in Table~\ref{tab:dataset_statistics} provide evidence to explain this observation; \textsc{IMDB-BINARY} and \textsc{COLLAB} have a significantly higher edge-to-node ratio when compared to the more sparse Cayley graph's structure. Nevertheless, our results reported in Table~\ref{tab:graph_rewiring} illustrate that the CGP model counteracts this by still providing leading performance on these datasets.

%% file: sections/appendices/virtual_nodes.tex
\begin{table}[t!]
    \scriptsize
    \setlength{\tabcolsep}{3pt}
    \centering
    \caption{Results of CGP using different virtual node initialisation strategies, including ZEROS, ONES, RANDOM and $\theta$ on the TUDataset. ZEROS is the baseline CGP and $\theta$ denotes the nodes are initialised using a learnable parameter.}
    \begin{tabular}{l cccccc}\toprule
        Model & \textsc{REDDIT-BINARY} & \textsc{IMDB-BINARY} & \textsc{MUTAG} & \textsc{ENZYMES} & \textsc{PROTEINS} & \textsc{COLLAB} \\ \midrule
        \;+ CGP-ZEROS & $78.225 \pm 1.268$ & $\textbf{71.650} \pm 1.532$ & $\textbf{85.250} \pm 3.200$ & $\textbf{50.083} \pm 2.242$ & $\textbf{73.080} \pm 1.396$ & $\textbf{73.350} \pm 0.788$ \\
        \;+ CGP-ONES & $79.175 \pm 1.992$ & $71.000 \pm 1.806$ & $82.250 \pm 3.773$ & $47.833 \pm 2.428$ & $70.535 \pm 1.972$ & $73.200 \pm 0.964$ \\
        \;+ CGP-RAND & $\textbf{82.300} \pm 1.368$ & $71.200 \pm 1.532$ & $84.000 \pm 3.130$ & $44.083 \pm 2.891$ & $70.000 \pm 1.937$ & $73.020 \pm 0.742$ \\
        \;+ CGP-$\theta$ & $80.150 \pm 1.888$ & $69.250 \pm 2.034$ & $85.000 \pm 3.606$ & $49.750 \pm 2.752$ & $70.133 \pm 2.053$ & $73.410 \pm 0.977$ \\
        \bottomrule
    \end{tabular}
    \label{tab:tu_virtual_nodes}
\end{table}

\begin{figure}[t!]
    \centering
    \subfigure[MUTAG]{\includegraphics[width=0.32\textwidth]{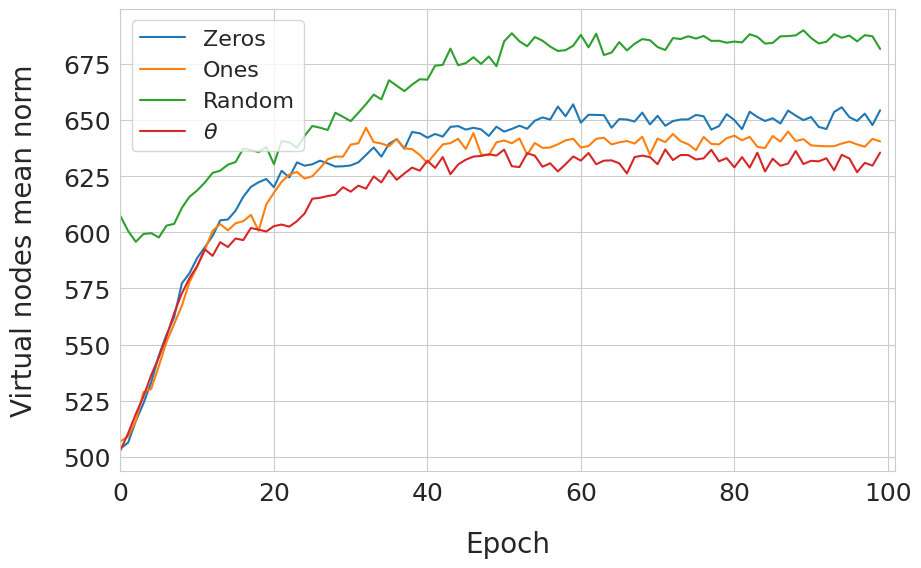}} 
    \subfigure[ENZYMES]{\includegraphics[width=0.32\textwidth]{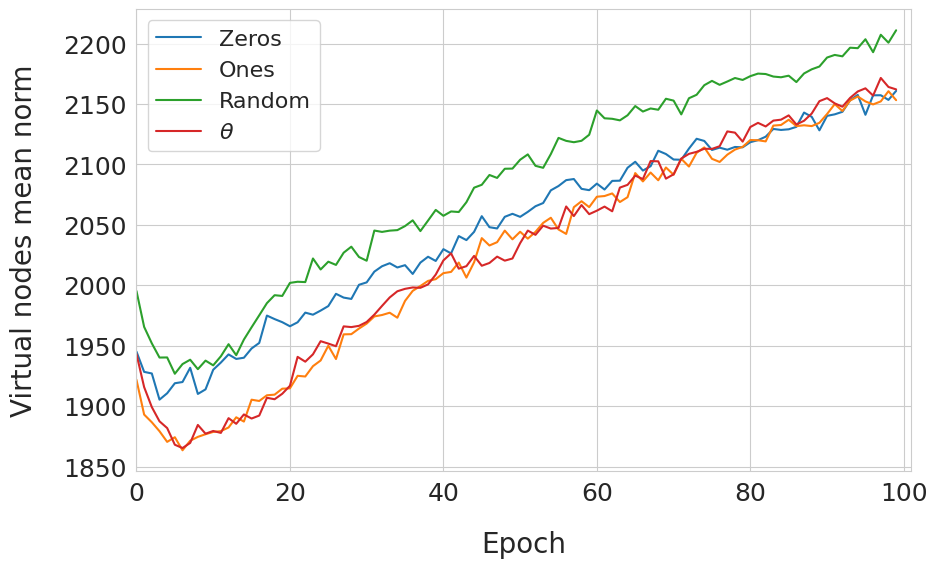}}
    \subfigure[PROTEINS]{\includegraphics[width=0.32\textwidth]{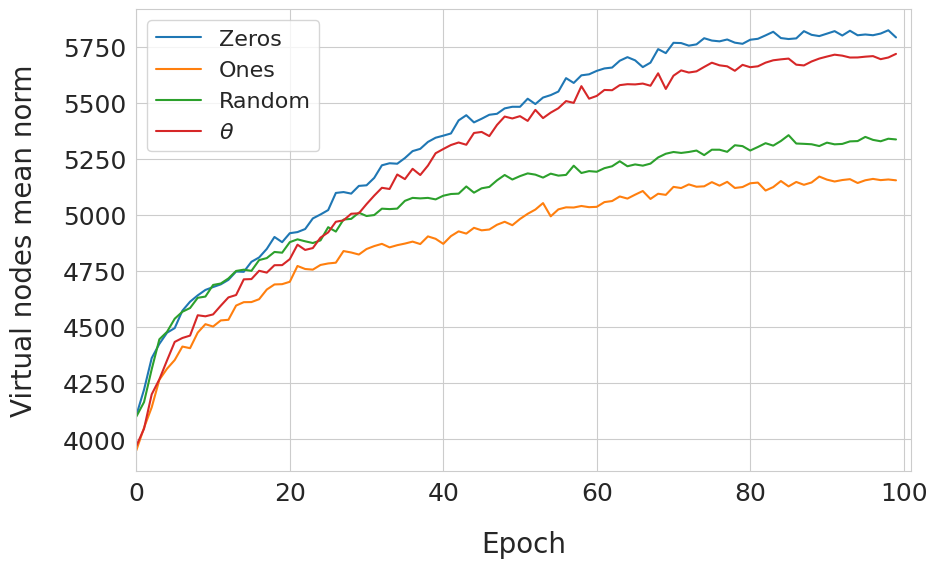}}
    \caption{The mean norm of the virtual node embeddings for CGP using different initialisation strategies on the TUDataset, including ZEROS, ONES, RANDOM and $\theta$.}
    \label{fig:norm}
\end{figure}

\begin{figure}[t!]
    \centering
    \subfigure[MUTAG]{\includegraphics[width=0.32\textwidth]{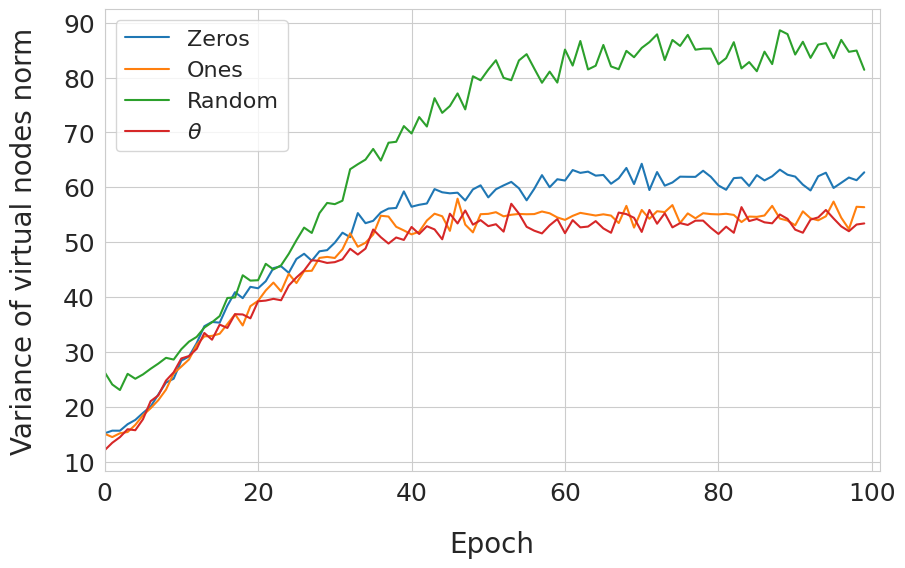}} 
    \subfigure[ENZYMES]{\includegraphics[width=0.32\textwidth]{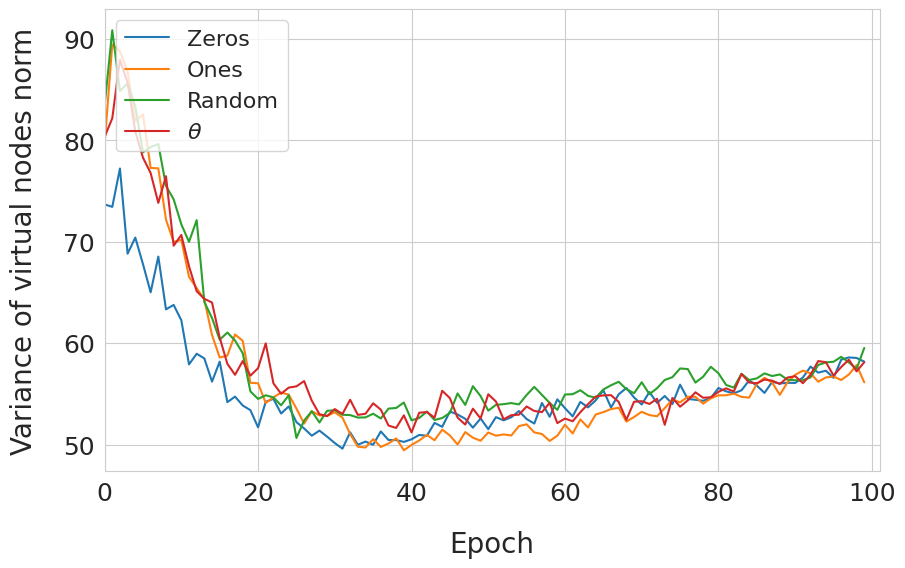}}
    \subfigure[PROTEINS]{\includegraphics[width=0.32\textwidth]{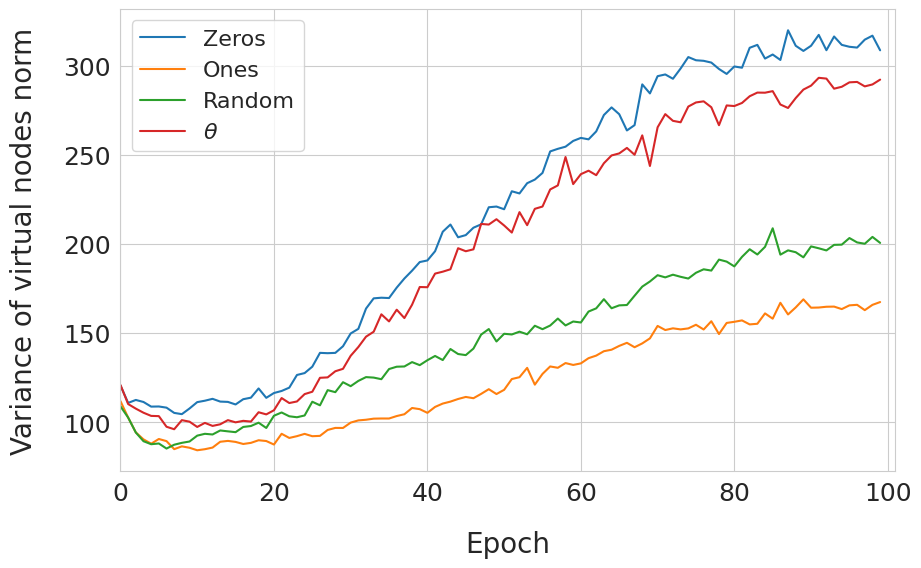}}
    \caption{The variance in the norm of the virtual node embeddings for CGP using different initialisation strategies on the TUDataset, including ZEROS, ONES, RANDOM and $\theta$.}
    \label{fig:var}
\end{figure}

In Section \ref{sec:cgp}, we show that virtual nodes in our proposed CGP method provide the flexibility to be initialised in some pre-defined manner. For the baseline CGP model, we opted to initialise them to zeros, which is in line with the work of \cite{gilmer2017neural, cai2023connection, southern2024understanding}. In this section, we empirically evaluate whether different node initialisation strategies impact the performance of CGP.  We report the results in Table \ref{tab:tu_virtual_nodes}, leveraging the TUDataset with the hyperparameter setting as prescribed in Section \ref{app:experimental_details}. For our other node initialisation strategies, we choose constant ones, random features that are sampled from $\mathcal{N}(0, 1)$ and $\theta$ where $\theta$ is a learnable parameter. Notably, ones is an interesting initialisation strategy as for datasets with no node features they are commonly assigned to a constant of ones \cite{morris2020tudataset}. The results overall show that our baseline model of CGP-ZERO performs the best across all datasets, however all approaches obtain similar performance within the variance of each other.

Next, we investigate the role of the additional virtual nodes in CGP. To this end, we use the mean norm of the virtual node embeddings, as well as the variance between them as a statistical proxy to measure the diversity of the embeddings. We visualise the results in Figure \ref{fig:norm} and \ref{fig:var} respectively for all of the virtual node initialisation strategies found in Table \ref{tab:tu_virtual_nodes}, however we fix the number of epochs to $100$ and report the average over 10 random seeds. The results show that during the training lifetime of the model, the mean norm and variance of virtual node node embeddings grow, indicating that they are learning distinct representations. Overall, the results in this section show that the initialisation of the virtual nodes impact CGP. However, due to the focus of CGP being to use complete Cayley graph structure and the additional virtual nodes serving as a conduit to this end. Therefore, a deeper understanding of the role of the sparsely connected virtual nodes and their initialisation strategies presents an interesting avenue for future work.

%% file: sections/appendices/experimental_details.tex
\begin{table}[t!]
    \scriptsize
    \setlength{\tabcolsep}{3pt}
    \centering
    \caption{Statistics of the datasets from OGB, TUDataset and LRGB used as part of our emperical evaluation. $^{\ast}$ denotes the number of additional virtual nodes (VN) added in our CGP model.}
    \begin{tabular}{l cccccc}\toprule
        Dataset & \#Graphs & Average \#Nodes & Average \#Edges & \#Classes & Average \#VN$^{\ast}$ & Max \#VN$^{\ast}$ \\ \midrule
        \textsc{OGBG-MOLHIV} & 41,127 & 25.5 & 27.5 & 2 & 11.8 & 191 \\
        \textsc{OGBG-PPA} & 158,100 & 243.4 & 2,266.1 & 37 & 50.3 & 191 \\ \midrule
        \textsc{REDDIT-BINARY} & 2,000 & 429.6 & 995.5 & 2 & 139.4 & 1809 \\
        \textsc{IMDB-BINARY} & 1,000 & 19.8 & 193.1 & 2 & 10.7 & 71 \\
        \textsc{MUTAG} & 188 & 17.9 & 39.6 & 2 & 8.0 & 23 \\
        \textsc{ENZYMES} & 600 & 32.6 & 124.3 & 6 & 14.6 & 71 \\
        \textsc{PROTEINS} & 1,113 & 39.1 & 145.6 & 2 & 20.6 & 190 \\
        \textsc{COLLAB} & 5,000 & 74.5 & 4,914.4 & 3 & 34.2 & 260 \\ \midrule
        \textsc{PEPTIDES-FUNC} & 15,535 & 150.94 & 307.3 & 10 & 67.0 & 263 \\
        \textsc{PEPTIDES-STRUCT} & 15,535 & 150.94 & 307.3 & 11 & 67.0 & 263 \\
        \bottomrule
    \end{tabular}
    \label{tab:dataset_statistics}
\end{table}

In this section, we provide thorough details regarding our experimental setting for each of our datasets. We utilise the well-established experimental settings \cite{hu2020open, karhadkar2023fosr} that are used for OGB and TUDataset respectively. The experimental setting for the LRGB \cite{dwivedi2022long} is found in Section~\ref{app:additional_experiments}. Refer to Table~\ref{tab:dataset_statistics} for the dataset statistics. Of significance, to rule out performance gain due to hyperparameter tuning in our results we use the same setting of GNN and hyperparameters for each task and model. All of our experiments use the default settings of the Adam optimiser \cite{kingma2015adam} with a learning rate of $1 \times 10^{-3}$, however the TUDataset and LRGB use the \textsc{ReduceLROnPlateau} scheduler with differing parameters.

\subsection{Datasets}

\paragraph{OGB Datasets}
From the OGB \cite{hu2020open} we consider two of the graph classification tasks: \textsc{ogbg-molhiv} and \textsc{ogbg-ppa}. \textsc{ogbg-molhiv} is among the largest molecule property prediction datasets within the scope of the MoleculeNet benchmark \citep{wu2018moleculenet}, and \textsc{ogbg-ppa} focuses on classifying species based on their taxa, using their protein-protein association networks \cite{proteins}. The OGB provides a unified evaluation protocol for each dataset, including application-specific data splits and evaluation metrics. Accordingly, our experimental setup to empirically evaluate against the OGB datasets leverages the \emph{official} open-source implementation from the OGB authors \cite{hu2020open}. \textsc{ogbg-molhiv} uses a 80\%/10\%/10\% train/validation/test split and ROC-AUC for the evaluation metric, whilst \textsc{ogbg-ppa} uses a 50\%/28\%/22\% train/validation/test split and accuracy for the evaluation metric. The hyperparameter setting for our models includes 5 layers, a hidden dimension of 300, a dropout of 50\% and the use of Batch Norm \cite{ioffe2015batch}. In Table~\ref{tab:ogb}, the results reported for our model are trained to 100 epochs across 10 seeds and 5 seeds for \textsc{ogbg-molhiv} and \textsc{ogbg-ppa} respectively.

\paragraph{TUDataset}
The TUDataset \cite{morris2020tudataset} is considered under the claim of \citet{karhadkar2023fosr} that the topology of the graphs in relation to the tasks require long-range interactions. Thus, we consider all graph classification tasks from the TUDataset: \textsc{REDDIT-BINARY}, \textsc{IMDB-BINARY}, \textsc{MUTAG}, \textsc{ENZYMES}, \textsc{PROTEINS} and \textsc{COLLAB}. Our setup for all TUDataset experiments is akin to the well-established and open-source setting of \citet{karhadkar2023fosr}. Accordingly, we train our GNNs with 80\%/10\%/10\% train/validation/test split and use a stopping patience of 100 epochs based on the validation loss. We fix the number of layers to 4 with a hidden dimension of 64 and a dropout of 50\%. The \textsc{ReduceLROnPlateu} uses the default setting as found in PyTorch~\cite{paszke2019pytorch}. Additionally, in accordance with~\citet{karhadkar2023fosr} the results for the TUDataset are reported to a 95\% confidence interval. However unlike \citet{karhadkar2023fosr}, we report the accuracy over 20 random seeds, set the maximum number of epochs to 300 \cite{choipanda}, as well as apply Batch Norm \cite{ioffe2015batch}. 

For TUDataset experiments, we compare CGP against the state-of-the-art approaches that require \emph{dedicated preprocessing} and use the hyperparameters as in the well-established baselines: DIGL~\cite{gasteiger2019digl}, SDRF~\cite{topping2022riccurvature}, FoSR~\cite{karhadkar2023fosr}, BORF~\cite{nguyen2023revisiting}, GTR~\cite{black2023understanding} and PANDA~\cite{choipanda}. By adopting the hyperparameters of these baselines methods, we not only ensure a fair comparison, but demonstrate that CGP remains competitive even under their optimised settings. For the graph rewiring techniques, we follow the hyperparameters as reported in the respective baselines. This includes the teleport probability ($\alpha$) and sparsification threshold ($\epsilon$) for DIGL, as well as the number of rewiring iterations for SDRF and FoSR being derived from~\citet{karhadkar2023fosr}. For BORF this includes the number of batches ($n$), number of edges added per batch ($h$), and number of edges removed per batch ($k$) from~\citet{nguyen2023revisiting}. For GTR this includes the number of edges added from \citet{black2023understanding}. Finally, this includes the hyperparameters for PANDA from \citet{choipanda}: the centrality metric $C(G)$, the top-$k$ nodes and the increased width (i.e. hidden dimension) of the model.

\subsection{Hardware}
\label{app:hardware}
All of our experimentation was conducted on a local machine with an AMD Ryzen 9 7950X3D 16-Core Processor (4.20 GHz), NVIDIA RTX 4090 ($24$ GB) and $64$ GB of RAM. The only exception is the \textsc{OGBG-PPA} results in which some of the baselines were processed on an external server with $8\times$ NVIDIA QUADRO RTX 8000 ($48$ GB). The following libraries were used as part of the implementation PyTorch~\cite{paszke2019pytorch}, PyTorch Geometric~\cite{fey2019fast} and NumPy~\cite{harris2020array}.

%% file: sections/appendices/additional_experiments.tex
In this section, we provide additional results to solidify the efficacy of our CGP model by comparing it against state-of-the-art graph rewiring techniques \cite{topping2022riccurvature, karhadkar2023fosr} and EGP \cite{deac2022expander} on the Long Range Graph Benchmark (LRGB) \cite{dwivedi2022long}. 

\begin{wraptable}{r}{8cm}
    \vspace{-5mm}
    \small
    \centering
    \caption{Comparative performance evaluation of CGP against graph rewiring techniques on the LRGB.}\label{tab:lrgb_rewiring}
    \begin{tabular}{l cccc}\toprule
        \multirow{2}{*}{Model} &\multicolumn{1}{c}{\textsc{Peptides-func}} &\multicolumn{1}{c}{\textsc{Peptides-struct}} \\ \cmidrule(lr){2-2} \cmidrule(lr){3-3} &\textbf{Test AP $\uparrow$} &\textbf{Test MAE $\downarrow$} \\ \midrule
        \midrule
        GCN & $0.5029 \pm 0.0058$ & $0.3587 \pm 0.0006$ \\
        \;+ SDRF & $0.5041 \pm 0.0026$  & $0.3559 \pm 0.0010$ \\
        \;+ FoSR & $0.4534 \pm 0.0090$ & $0.3003 \pm 0.0007$ \\
        \;+ EGP & $0.4972 \pm 0.0023$ & $0.3001 \pm 0.0013$ \\
        \midrule
        \;+ \textbf{CGP} & $\textbf{0.5106} \pm 0.0014$ & $\textbf{0.2931} \pm 0.0006$ \\
        \midrule \midrule
        GIN & $0.5124 \pm 0.0055$ & $0.3544 \pm 0.0014$ \\
        \;+ SDRF & $0.5122 \pm 0.0061$  & $0.3515 \pm 0.0011$ \\
        \;+ FoSR & $0.4584 \pm 0.0079$ & $0.3008 \pm 0.0014$ \\
        \;+ EGP & $0.4926 \pm 0.0070$ & $0.3034 \pm 0.0027$ \\
        \midrule
        \;+ \textbf{CGP} & $\textbf{0.5159} \pm 0.0059$ & $\textbf{0.2910} \pm 0.0011$ \\
        \bottomrule
    \end{tabular}
\end{wraptable}

\paragraph{Datasets}
We consider the \textsc{Peptides} datasets from the Long Range Graph Benchmark (LRGB) \cite{dwivedi2022long}, which have two related tasks \textsc{Peptides-func} and \textsc{Peptides-struct}. The former is a peptide feature classification task in which the objective is to predict the peptide function out of 10 classes with the performance being measured by Average Precision (AP). The latter consists of the same graphs as \textsc{Peptides-func}, however instead it is a graph regression task in which the aim is to predict aggregated 3D properties of the peptides at the graph level; the performance metric is Mean Absolute Error (MAE). The dataset statistics for the LRGB can be found in Table~\ref{tab:dataset_statistics}.

\paragraph{Experimental details}
Similar to the OGB, the LRGB \cite{dwivedi2022long} provides an experimental setting with the aim to have unified experimental evaluation of their benchmarks. Correspondingly, we leverage the LRGB implementation that is built upon the GraphGym module \cite{you2020design}. For both \textsc{Peptides} tasks it uses a 70\%/15\%/15\% train/validation/test split. The experimental setup of \citet{dwivedi2022long} fixes the models number of layers to 5, and does not use dropout, but it does use Batch Norm \cite{ioffe2015batch}. However, we reduce the models number of parameters using a hidden dimension of 64 as in \citet{nguyen2023revisiting}, as opposed to 300 \cite{dwivedi2022long}. Additionally, we reduce the number of epochs to 250, which is in line with \citet{tonshoff2023did}. A \textsc{ReduceLROnPlateau} scheduler is used following \citet{dwivedi2022long} settings of a patience of 20 epochs, a decay factor of 0.5 and a minimum learning rate of $1 \times 10^{-5}$. We use the graph rewiring hyperparameters from \cite{nguyen2023revisiting, barbero2023locality}. 

\paragraph{Results}
The results in Table~\ref{tab:lrgb_rewiring} showcase that CGP outperforms the state-of-the-art graph rewiring approaches, as well as EGP for both GCN and GIN. However, it is noted that the work of \citet{tonshoff2023did} achieves improved performance through extensive hyperparameter tuning. Nevertheless, this is beyond the scope of evaluating the impact of using the complete Cayley graph structure.

%% file: sections/appendices/scalability.tex
\begin{figure}[t!]
    \centering
    \includegraphics[width=0.48\textwidth]{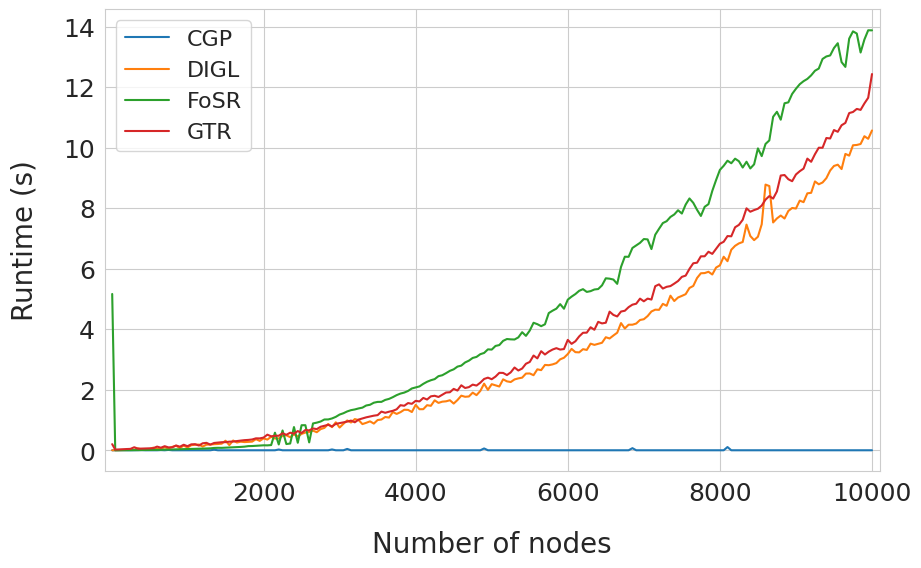}
    \includegraphics[width=0.48\textwidth]{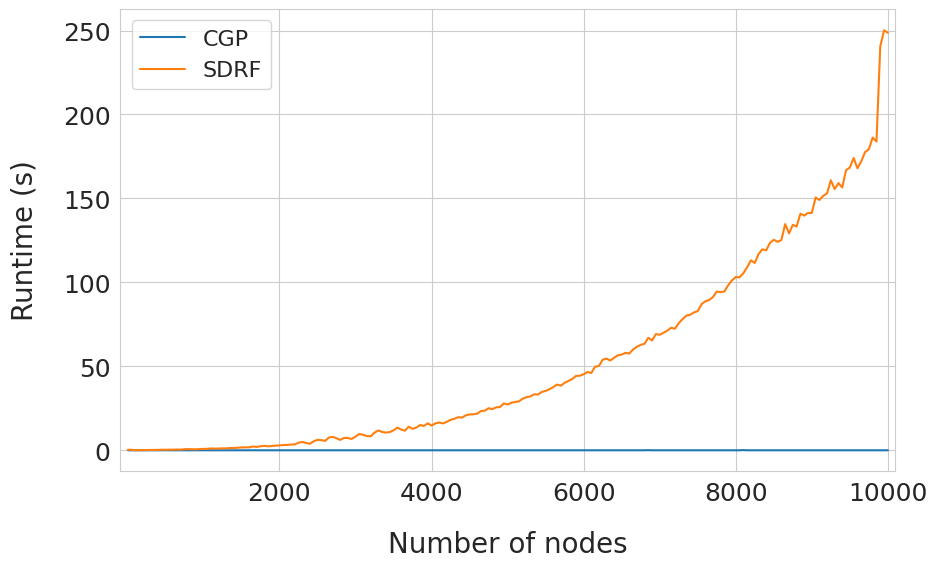}
    \caption{Synthetic preprocessing benchmark for CGP in regards to graph rewiring techniques, using Erdős–Rényi graphs with a probability $p = \frac{5 \log n}{n}$. \textbf{Left:} Preprocessing time of CGP against DIGL, FoSR and GTR. \textbf{Right:} Preprocessing time of CGP against SDRF.}
    \label{fig:scalability}
\end{figure}

\begin{table}[ht!]
    \small
    \setlength{\tabcolsep}{3pt}
    \centering
    \caption{Preprocess graph rewiring runtime (in seconds) for each graph in the TUDataset. OOT indicates out-of-time for the \emph{prepreprocessing rewiring} time.}
    \begin{tabular}{l cccccc}\toprule
        Model & \textsc{REDDIT-BINARY} & \textsc{IMDB-BINARY} & \textsc{MUTAG} & \textsc{ENZYMES} & \textsc{PROTEINS} & \textsc{COLLAB} \\ \midrule
        DIGL & 40.3837 & 0.411771 & 0.0342833 & 0.243485 & 0.491458 & 56.3175 \\ 
        SDRF & 359.128 & 5.13257 & 0.669701 & 1.71482 & 3.02873 & 619.125 \\ 
        FoSR & 74.8568 & 4.54634 & 4.71567 & 4.56855 & 5.04358 & 9.79994 \\
        BORF & OOT & 465.408 & 53.7069 & 179.573 & 351.173 & OOT \\ 
        GTR & 118.549 & 3.39839 & 1.54127 & 2.87399 & 6.49714 & 92.6125 \\
        PANDA & 6.13925 & 0.789759 & 0.246243 & 0.278594 & 0.248043 & 230.850 \\
        \midrule
        EGP & 0.245215 & 0.0185697 & 0.00446963 & 0.0163198 & 0.0393348 & 0.129567\\
        CGP & 0.226065 & 0.0211341 & 0.00438905 & 0.0166841 & 0.0348585 & 0.131188 \\
        \bottomrule
    \end{tabular}
    \label{tab:graph_rewiring_times_tu}
\end{table}

\begin{table}[ht!]
    \small
    \setlength{\tabcolsep}{3pt}
    \centering
    \caption{Preprocess runtime (in seconds) for state-of-the-art graph rewiring techniques for each graph in the LRGB dataset.}
    \begin{tabular}{l cccccc}\toprule
        Model & \textsc{Peptides-func} & \textsc{Peptides-struct} \\ \midrule
        SDRF & 61.0356 & 56.1561 \\ 
        FoSR & 23.4263 & 24.1858 \\
        \midrule
        EGP & 1.36170 & 1.29376 \\
        CGP & 1.27776 & 1.29608 \\
        \bottomrule
    \end{tabular}
    \label{tab:graph_rewiring_times_lrgb}
\end{table}

We empirically analyse the scalability of CGP by comparing the computational preprocessing time against the state-of-the-art graph rewiring techniques~\cite{gasteiger2019digl, topping2022riccurvature, karhadkar2023fosr, nguyen2023revisiting, black2023understanding}, as well as PANDA \cite{choipanda}. We first provide a real-world evaluation by benchmarking the preprocessing time on two real-world datasets: TUDataset~\cite{morris2020tudataset} and LRGB~\cite{dwivedi2022long}. The results are reported in Table~\ref{tab:graph_rewiring_times_tu} and~\ref{tab:graph_rewiring_times_lrgb}, using the same graph rewiring techniques as in Table~\ref{tab:graph_rewiring} and~\ref{tab:lrgb_rewiring} respectively. To extend our scalability analysis, we create a synthetic benchmark by leveraging Erdős–Rényi with a probability $p = \frac{5 \log n}{n}$ (as used by \citet{karhadkar2023fosr}) to create graphs of up to 10,000 nodes. In line with the results reported in Table~\ref{tab:graph_rewiring}, we do not conduct the synthetic benchmark for BORF \cite{nguyen2023revisiting}, due to the impracticality of the rewiring time which is highlighted in Section \ref{sec:experimentation}. 

The results show the efficacy of our proposed CGP model, as the preprocessing time is orders of magnitude lower than the graph rewiring techniques. To this end, we examine the lack of computational preprocessing time required to generate the corresponding Cayley graphs for both CGP and EGP. Overall, our results show the practicality of CGP to scale to large graphs when compared to the graph rewiring approaches. This is reinforced by the experimentation being conducted on the local machine with leading hardware as in Section \ref{app:hardware}. In particular, both the CPU and GPU deliver top-tier clock speeds, therefore on lower-performing hardware, the graph rewiring techniques could have a more detrimental effect.

%% file: sections/appendices/regular_graphs.tex
\begin{figure*}[t]
\centering
        \includegraphics[width=0.5\linewidth]{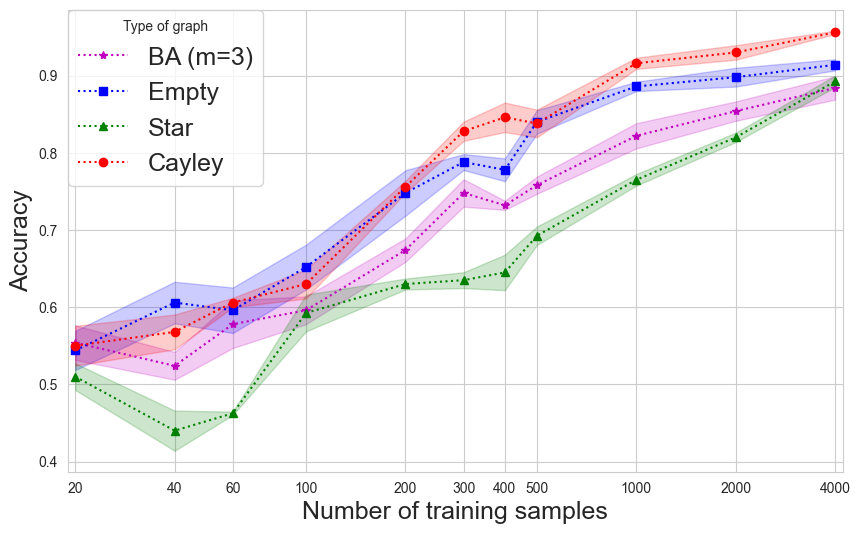}
        \caption{The learning curves of the same GNN model trained on graphs that have the same node features and only differ in their graph structure, which is sampled from different distributions.  
    The label is computed from the node features without the use of any graph structure. The GNN overfits the graph structure instead of ignoring it, and therefore the model performance differ across different graph distributions. Cayley graphs exhibit the best performance, and robustness to overfitting.}
        \label{Figure:cayley_exp}
\end{figure*}

In this section, we explore the additional benefits of Cayley graphs being a regular graph. It was observed by \citet{bechler2023graph} that GNNs tend to overfit the given graph structure, even in cases where it does no provide useful information for the predictive task. Nonetheless, it was shown that regular graphs exhibit robustness to this overfitting. As Cayley graphs are regular graphs, they exhibit this robustness.

We repeat the experiments from \cite{bechler2023graph} to ensure the robustness of Cayley graphs to graph overfitting. 
The task is a binary classification task where the label is independent of the graph, and is computed only over the features. 
We used the Sum task that was presented in \cite{bechler2023graph}: 
the label is generated using a teacher GNN that simply sums the node features and applies a linear readout to produce a scalar. 

We used four different datasets from this baseline by sampling graph-structures from different graph distributions. The set of node feature vectors remains the same across all the datasets, and thus, the datasets differ only in their graph structure.
The graph distributions we used are: Cayley graphs over $24$ nodes,  star-graph (Star) where the only connections are between one specific node and all other nodes, and the preferential attachment model (BA) ~\cite{badist}, where the graph is built by incrementally adding new nodes and connecting them to existing nodes with probability proportional to the degrees of the existing nodes. We used the data as is with empty graphs (Empty) as a baseline to compare to. On each dataset, we varied the training set size and evaluated test errors on $5$ runs with random seeds.

The results are shown in Figure~\ref{Figure:cayley_exp}.
The GNN trained on the Cayley graphs performs similarly to when trained on empty graphs. Nonetheless, when trained on other distributions, the performance decreases and does not recover even with $4000$ training samples. This demonstrate the robustness of Cayley graphs to graph overfitting.

\paragraph{Extrapolation}
The ability and failures of GNNs to extrapolate to graphs of sizes larger then the one presented during training was examined in~\citet{yehudai2021localstructuressizegeneralization}. It was shown that size generalisation is dependent on local structures around each node, called \textit{d-patterns}. In particular, if increasing the graph size does not change the distribution of these d-patterns, then extrapolation to larger graph sizes is guaranteed.

\paragraph{Sum Task}
This is a binary classification synthetic task with a graph-less ground truth function. To generate the label, we use a teacher GNN that simply sums the node features and applies a linear readout to produce a scalar.
The data contains non-informative graph-structures which are drawn from the GNP graph distribution~\cite{gnp}, where the edges are sampled i.i.d with probability $p$ (we used $p=0.5$). 

The teacher readout is sampled once from $\mathcal{N}(0,1)$ and used for all the graphs. All graphs have $n=20$ nodes, and each node is assigned with a feature vector in $\mathbb{R}^{128}$ sampled i.i.d from $\mathcal{N}(0, 1)$. We used a $1$-layer ``student" GNN following the teacher model, with readout and ReLU activations. 

We evaluated the learning curve with an increasing amount of $\{20, 40, 60, 100, 200, 300, 400, 500, 1000, 2000, 4000\}$ samples. We note that the GNN has a total of $\sim$16,000 parameters, and thus, it is over-parameterised and can fit the training data with perfect accuracy.

%% file: sections/appendices/dirichlet_energy.tex
\begin{figure}[t!]
    \centering
    \subfigure[MUTAG]{\includegraphics[width=0.32\textwidth]{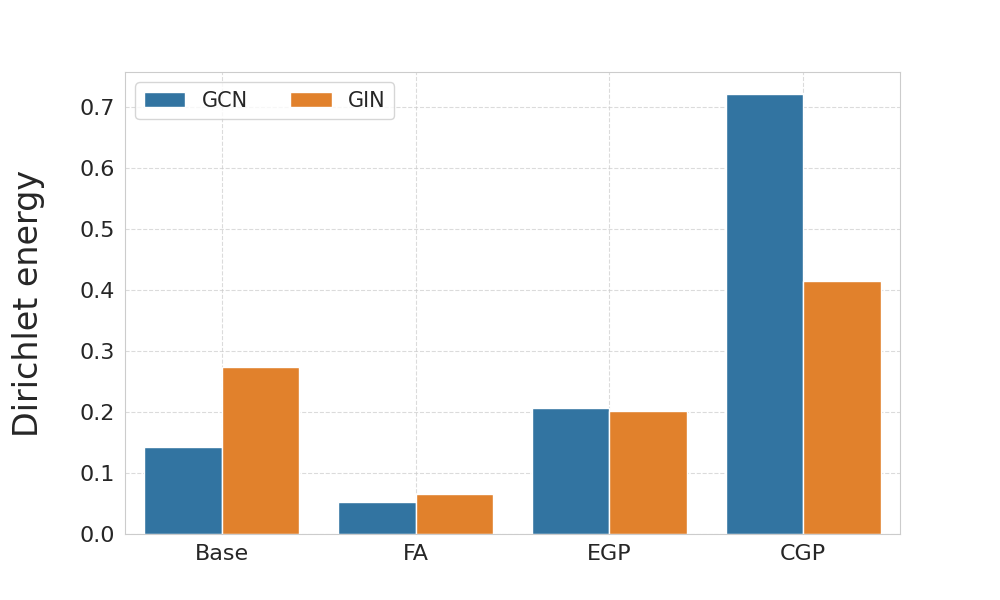}} 
    \subfigure[ENZYMES]{\includegraphics[width=0.32\textwidth]{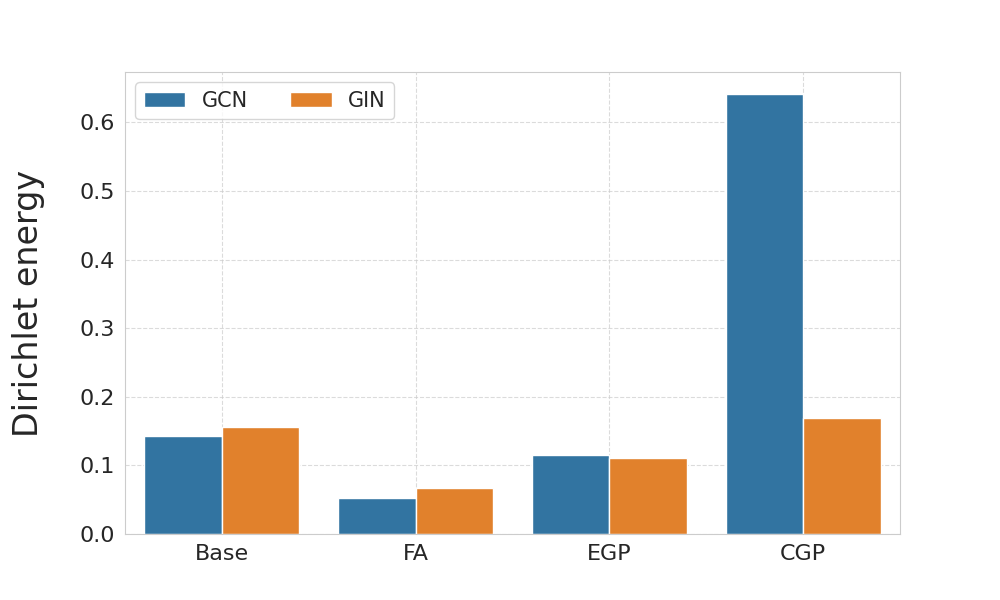}}
    \subfigure[PROTEINS]{\includegraphics[width=0.32\textwidth]{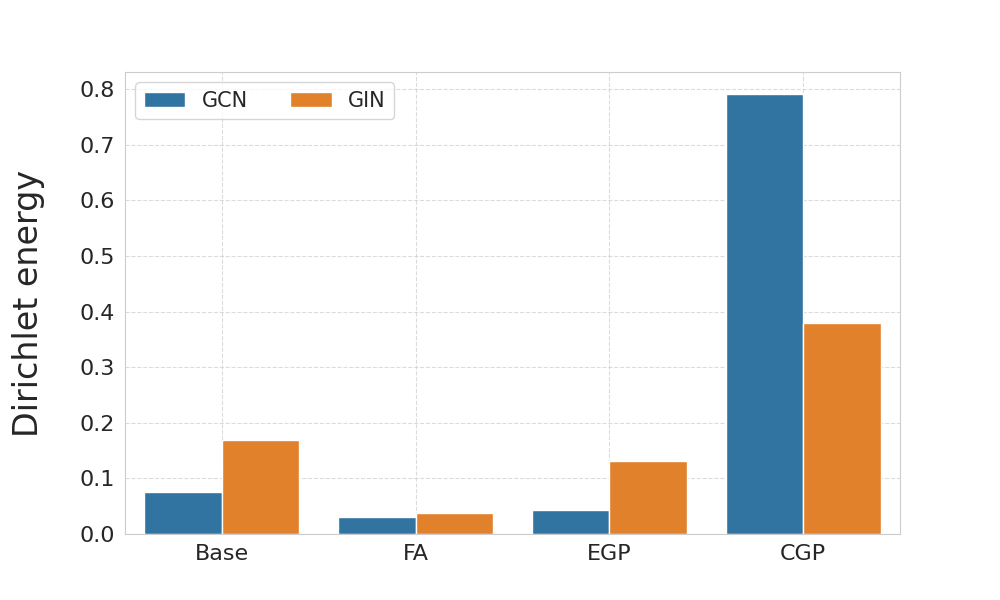}}
    \caption{Comparison of the Dirichlet energy for CGP against the baseline model, FA and EGP for the TUDataset. A higher energy indicates that the proposed approach is more robust to the over-smoothing problem.}
    \label{fig:dirichlet_energies}
\end{figure}

Here, we evaluate the impact of propagating over the complete Cayley graph structure in regards to the over-smoothing problem using the Dirichlet energy. Over-smoothing is an independent problem from over-squashing, but another well-known problem that impacts the expressivity of GNNs \cite{nt2019revisiting, rusch2023survey}. This phenomenon occurs in GNNs when the number of layers increases \cite{li2018deeper, oono2019graph}, such that node features become increasingly similar \cite{di2022understandingenergies}. However, the over-smoothing problem is linked with over-squashing due to a common approach to the latter being graph rewiring; too many additional edges lead to over-smoothing \cite{karhadkar2023fosr}. There are varying approaches to measure over-smoothing for a graph with one such notable metric being the Dirichlet energy \cite{rusch2022graph, karhadkar2023fosr, arnaiz2022diffwire}.

The Dirichlet energy quantifies over-smoothing by measuring the deviation of a function on the graph from being constant between connected node pairs, thus indicating the level of non-smoothness in the signals \cite{chungbook}. In turn, the Dirichlet energy has been used to measure the amount of over-smoothing in graph representations~\cite{karhadkar2023fosr, arnaiz2022diffwire, choipanda}.

Similar to EGP~\cite{deac2022expander} and FA~\cite{alon2020bottleneck}, the CGP model uses an independent graph structure to propagate information over, as opposed to the graph rewiring approaches in which they directly alter the input graph structure. Consequently, we conduct our Dirichlet energy analysis against EGP and FA, which also fall under the category of approaches that do not require \emph{dedicated preprocessing}. The results in Figure~\ref{fig:dirichlet_energies} show that CGP consistently obtains a higher Dirichlet energy for both GCN and GIN when compared to EGP and FA. This further highlights the strengths of the CGP model to mitigate over-squashing, whilst minimising the negative implications of over-smoothing through the use of the \emph{additional virtual nodes} retained from the complete Cayley graph structure.